\documentclass[preprint,sort&compress,12pt]{elsarticle}

\usepackage{amssymb}
\usepackage{amsthm}
\usepackage{amsmath}
\usepackage{mathtools}
\usepackage{mathrsfs}
\usepackage{algorithm}
\usepackage[algo2e]{algorithm2e}
\usepackage{algpseudocode}
\usepackage{array}
\usepackage{multirow}
\usepackage{listings}
\usepackage{tabu}
\usepackage{enumerate}
\usepackage{lineno}
\usepackage{fullpage}
\usepackage{xcolor}
\usepackage[colorinlistoftodos]{todonotes}
\usepackage{bbm}
\usepackage[colorlinks=true]{hyperref}
\usepackage{url}
\usepackage{textcomp}
\usepackage{gensymb}
\usepackage{soul}
\usepackage{graphicx}
\usepackage{subfig}
\usepackage{float}
\usepackage{caption}
\usepackage{tikz}
\usetikzlibrary{shapes}

\usepackage{wrapfig}
\usepackage{enumitem}
\usepackage{nomencl}

\theoremstyle{definition}

\theoremstyle{remark}

\biboptions{numbers,comma,round,square}
\graphicspath{ {./fig/} }

\journal{Elsevier}
\begin{document}
\makeatletter
\def\ps@pprintTitle{%
  \let\@oddhead\@empty
  \let\@evenhead\@empty
  \let\@oddfoot\@empty
  \let\@evenfoot\@oddfoot
}

\begin{frontmatter}

\title{Implicit Neural Differential Model for Spatiotemporal Dynamics}

\author[ndAME]{Deepak Akhare}
\author[ndAME]{Pan Du}
\author[ndAME]{Tengfei Luo\corref{corjw}}
\author[ndAME,cornell]{Jian-Xun Wang\corref{corjw}}


\address[ndAME]{Department of Aerospace and Mechanical Engineering, University of Notre Dame, Notre Dame, IN, USA}
\address[cornell]{Sibley School of Mechanical and Aerospace Engineering, Cornell University, Ithaca, NY, USA}
\cortext[corjw]{Corresponding author: jw2837@cornell.edu, tluo@nd.edu}

\begin{abstract}

Hybrid neural–physics modeling frameworks through differentiable programming have emerged as powerful tools in scientific machine learning, enabling the integration of known physics with data-driven learning to improve prediction accuracy and generalizability. However, most existing hybrid frameworks rely on explicit recurrent formulations, which suffer from numerical instability and error accumulation during long-horizon forecasting. In this work, we introduce Im-PiNDiff, a novel implicit physics-integrated neural differentiable solver for stable and accurate modeling of spatiotemporal dynamics. Inspired by deep equilibrium models, Im-PiNDiff advances the state using implicit fixed-point layers, enabling robust long-term simulation while remaining fully end-to-end differentiable.
To enable scalable training, we introduce a hybrid gradient propagation strategy that integrates adjoint-state methods with reverse-mode automatic differentiation. This approach eliminates the need to store intermediate solver states and decouples memory complexity from the number of solver iterations, significantly reducing training overhead. We further incorporate checkpointing techniques to manage memory in long-horizon rollouts. Numerical experiments on various spatiotemporal PDE systems, including advection–diffusion processes, Burgers’ dynamics, and multi-physics chemical vapor infiltration processes, demonstrate that Im-PiNDiff achieves superior predictive performance, enhanced numerical stability, and substantial reductions in memory and runtime cost relative to explicit and naive implicit baselines. This work provides a principled, efficient, and scalable framework for hybrid neural–physics modeling.

\end{abstract}


\end{frontmatter}

\section{Introduction}
\label{sect:introduction}

Computational science is experiencing a transformative shift, driven by advancements in numerical techniques, artificial intelligence (AI), and the growing availability of extensive datasets. At the core of this transformation lies the scientific machine learning (SciML), a rapidly evolving field that synergistically integrates physics-based modeling with modern machine learning (ML). By combining data-driven insights with fundamental physical principles, SciML offers unprecedented opportunities to accurately model complex systems, enhance predictive capabilities, and optimize computational workflows across a wide range of scientific and engineering applications.

Notable methodologies underpinning SciML include physics-informed neural networks (PINNs)~\cite{raissi2019physics,sun2020surrogate,sun2020physics}, neural operators~\cite{li2021fourier,lu2021learning,wang2021learning}, equation discovery techniques~\cite{brunton2016discovering,chen2021physics,sun2022bayesian}, and hybrid neural-physics models~\cite{kochkov2021machine,liu2024multi,fan2023differentiable,akhare2023physics,akhare2024probabilistic,akhare2023diffhybrid,fan2024neural}. 
Among these, hybrid neural-physics models have attracted significant attention due to their explicit incorporation of domain knowledge into learning architecture, addressing the key limitations of purely data-driven approaches, such as limited extrapolation and generalization capabilities. This integration ensures consistency with known physical laws meanwhile enabling us to capture uncharacterized system dynamics, thus balancing flexibility with reliability. Historically, however, most hybrid frameworks have employed weak coupling strategies, wherein ML models are trained offline and subsequently embedded into conventional numerical solvers~\cite{tompson2017accelerating,vinuesa2022enhancing,margenberg2024dnn}. Such approaches, while prevalent in turbulence modeling~\cite{wang2017physics,duraisamy2019turbulence,wang2019prediction} and atmospheric simulations~\cite{zanna2020data,mendez2023data}, are fundamentally limited. Their reliance on separately trained components prevents end-to-end optimization, restricting their robustness, generalization, and adaptability to complex, unseen scenarios.

Recognizing these challenges, recent trends advocate strongly integrated frameworks that enable end-to-end differentiable hybridization, leveraging indirect and sparse observational data. This integration is made possible by differentiable programming (DP)~\cite{newbury2024review}, which facilitates joint optimization of ML models and numerical solvers. Recent advances in differentiable physics and hybrid neural-physics models have demonstrated significant potential across various scientific domains~\cite{list2025differentiability,schweidtmann2023review,innes2019differentiable,belbute2020combining,huang2020learning,kochkov2021machine,list2022learned,akhare2023physics,liu2024multi,fan2023differentiable,fan2024differentiable}. For example, Kochkov et al.~\cite{kochkov2021machine} utilized convolutional neural networks (CNNs) to accelerate differentiable computational fluid dynamics (CFD) solvers, while Huang et al.~\cite{huang2020learning} embedded neural networks within a differentiable finite element solver to learn constitutive relations from indirect observations. Further advances by Wang and coworkers introduced a physics-integrated neural differentiable (PiNDiff) modeling framework, unifying neural operators with numerical PDE solvers to achieve enhanced generalization and accuracy~\cite{akhare2023physics,liu2024multi,fan2023differentiable, fan2024differentiable,fan2024neural}. 

Despite their considerable potential, current PiNDiff frameworks predominantly utilize explicit, auto-regressive recurrent architectures. Such explicit recurrent structures suffer from inherent numerical instability, error accumulation, and deteriorating performance in long-term predictions, particularly for stiff or chaotic systems, thereby limiting their practical applicability. Inspired by classical numerical analysis, where implicit methods offer superior stability properties, this paper proposes an innovative implicit neural differential model (Im-PiNDiff) for robust spatiotemporal dynamics prediction. By employing implicit neural network layers, our framework mitigates error accumulation and significantly enhances numerical stability and accuracy, enabling reliable long-term simulations.

However, adopting implicit neural architectures within differentiable frameworks introduces considerable computational and memory challenges. This is primarily due to the requirement of automatic differentiation (AD), which involves storing intermediate activations, computational graphs, and input data as buffers during forward propagation to compute gradients~\cite{margossian2019review,baydin2018automatic}.  These memory requirements grow exponentially in implicit learning architectures, involving bilevel optimization, iterative solvers, and extended simulations, frequently leading to prohibitive training times~\cite{blondel2024elements,fan2024differentiable,newbury2024review}. To address these computational hurdles, we introduce a hybrid training strategy that combines adjoint state methods with reverse-mode AD. The adjoint method decouple memory requirements from iterative solver iterations, significantly reducing computational overhead and memory usage.  Our approach employs adjoint-based methods to compute and propagate gradients over the implicit layers while utilizing reverse-mode AD for explicit model components. Further, we employ strategic checkpointing techniques to optimize memory usage, ensuring scalability and practical feasibility for large-scale, complex problems. In summary, the key contributions of this work include: (1) a novel implicit PiNDiff framework integrating implicit neural architectures, differentiable numerical PDEs, and conditional neural field, enabling stable and accurate long-term predictions of complex spatiotemporal dynamics; (2) a hybrid gradient computation strategy that leverages adjoint-state methods and reverse-mode AD, significantly improving computational efficiency and reducing memory overhead; (3) numerical validation demonstrating improved performance, stability, and computational feasibility of the proposed Im-PiNDiff framework. Together, these innovations represent a significant step toward enabling the efficient and scalable application of implicit PiNDiff frameworks.

The rest of this paper is organized as follows: Section~\ref{sec:methodology} details the proposed methodology, including the mathematical formulation, implicit neural differentiable model, and hybrid training strategies. Section~\ref{sec:results} presents numerical experiments demonstrating the framework’s performance across a range of applications. Finally, Section~\ref{sec:conclusion} summarizes the contributions and outlines potential future research directions.


\section{Methodology}
\label{sec:methodology}

\subsection{Problem formulation}
\label{sec:problem}

Most fundamental physical laws governing phenomena across diverse scientific and engineering disciplines, such as fluid dynamics, solid mechanics, heat transfer, electromagnetism, and quantum mechanics, are naturally expressed in the mathematical form of partial differential equations (PDEs). In practice, however, the exact forms of these PDEs often contain unknown or uncertain components due to incomplete understanding of underlying physics or inherent modeling limitations. Such scenarios can be represented by generic PDEs,
\begin{linenomath*}
\begin{subequations}
    \begin{alignat}{3}
    \frac{\partial \phi}{\partial t} = \mathscr{F}\big[\mathcal{K} (\phi(\tilde{\mathbf{x}}); \boldsymbol{\lambda}_{\mathcal{K}}, \boldsymbol{\lambda}_{\mathcal{U}}), \mathcal{U} (\phi(\tilde{\mathbf{x}}); \boldsymbol{\lambda}_{\mathcal{K}}, \boldsymbol{\lambda}_{\mathcal{U}})\big], \ \ \ \ &\tilde{\mathbf{x}} \in  \Omega_{p,t}
    \label{eq:PiNDiff-PDE},\\
    \mathcal{BC}(\phi(\tilde{\mathbf{x}}); \boldsymbol{\lambda}_{\mathcal{K}}, \boldsymbol{\lambda}_{\mathcal{U}}) = 0,\ \  \ \ &\tilde{\mathbf{x}} \in  \partial\Omega_{p,t}
    \label{eq:PiNDiff-BC},\\
    \mathcal{IC}(\phi(\tilde{\mathbf{x}}); \boldsymbol{\lambda}_{\mathcal{K}}, \boldsymbol{\lambda}_{\mathcal{U}}) = 0,\ \  \ \ &\tilde{\mathbf{x}} \in  \Omega_{p, t=0}
    \label{eq:PiNDiff-IC},
    \end{alignat}
    \label{eq:PiNDiff}
\end{subequations}
\end{linenomath*}
where nonlinear functions $\mathcal{K}(\cdot)$ and $\mathcal{U}(\cdot)$ represent the known and unknown components of the PDEs for state variable $\phi$, coupled via the nonlinear functional $\mathscr{F}(\cdot)$. The initial and boundary conditions are abstractly defined by the PDE operators $\mathcal{IC}$ and $\mathcal{BC}$, respectively.
These functions rely on a set of physical parameters, with the known ones denoted by $\boldsymbol{\lambda}_\mathcal{K}$ and uncertain ones by $\boldsymbol{\lambda}_\mathcal{U}$, respectively. The spatial and temporal coordinates are denoted as $\tilde{\mathbf{x}} = \{\textbf{x}, t\}$, whit physical domain $\Omega_{p}$, boundary $\partial\Omega_{p}$, and time domain $[0, T]$, resulting in a spatiotemporal domain $\Omega_{p,t} \triangleq \Omega_{p} \times [0, T]$.

Due to incomplete PDE formulations, hybrid neural models based on DP, e.g., PiNDiff, integrate deep neural networks (DNNs) to approximate unknown PDE components/operators or enhance known components through learnable operators. The integrated neural network parameters $\boldsymbol{\theta}_{nn}$ and uncertain physical parameters $\boldsymbol{\lambda}_{\mathcal{U}}$ form a unified set of trainable parameters $\boldsymbol{\theta} = [\boldsymbol{\theta}_{nn}, \boldsymbol{\lambda}_{\mathcal{U}}]^T$, optimized concurrently as part of a unified network, enabled by DP. The training of PiNDiff model over dataset $\mathcal{D}=\{\tilde{\mathbf{x}}_i,{\phi}_i\}^N_{i=1}$  is formulated as a PDE-constrained optimization problem,
\begin{subequations}\label{eq:Hybrid_opt}
\begin{align}
\min_{\boldsymbol{\theta}} \quad & J(\phi_{\boldsymbol{\theta}}(\tilde{\mathbf{x}}), \mathcal{D}; \boldsymbol{\theta}) \\
\textrm{s.t.} \quad & 
    \frac{\partial \phi_{\boldsymbol{\theta}}(\tilde{\mathbf{x}})}{\partial t} = \mathscr{F}_{nn}\big[\mathcal{K}(\phi_{\boldsymbol{\theta}}(\tilde{\mathbf{x}}); \boldsymbol{\lambda}_{\mathcal{K}}, \boldsymbol{\lambda}_{\mathcal{U}}), \mathcal{U}_{nn}(\phi_{\boldsymbol{\theta}}(\tilde{\mathbf{x}}); \boldsymbol{\lambda}_{\mathcal{K}}, \boldsymbol{\lambda}_{\mathcal{U}},\boldsymbol{\theta}_{nn}); \boldsymbol{\theta}_{nn}\big], \ \ \ \ \tilde{\mathbf{x}} \in  \Omega_{p,t} \label{eq:Hybrid_opt_a}\\
  &
    \mathcal{BC}(\phi_{\boldsymbol{\theta}}(\tilde{\mathbf{x}}); \boldsymbol{\lambda}_{\mathcal{K}}, \boldsymbol{\lambda}_{\mathcal{U}}) = 0,\ \  \ \ \tilde{\mathbf{x}} \in  \partial\Omega_{p,t} \label{eq:Hybrid_opt_b}\\
  &
    \mathcal{IC}(\phi_{\boldsymbol{\theta}}(\tilde{\mathbf{x}}); \boldsymbol{\lambda}_{\mathcal{K}}, \boldsymbol{\lambda}_{\mathcal{U}}) = 0,\ \  \ \ \tilde{\mathbf{x}} \in  \Omega_{p, t=0} \label{eq:Hybrid_opt_c}
\end{align}
\end{subequations}
where $J$ is the objective function. Typically, existing PiNDiff models rely on explicit recurrent formulations, wherein solutions at each time step explicitly depend on the previous state. Although straightforward, these auto-regressive recurrent approaches commonly suffer from instability and error accumulation, especially during long-horizon predictions. 

\begin{figure}[!htp]
    \centering
    \includegraphics[width=\linewidth]{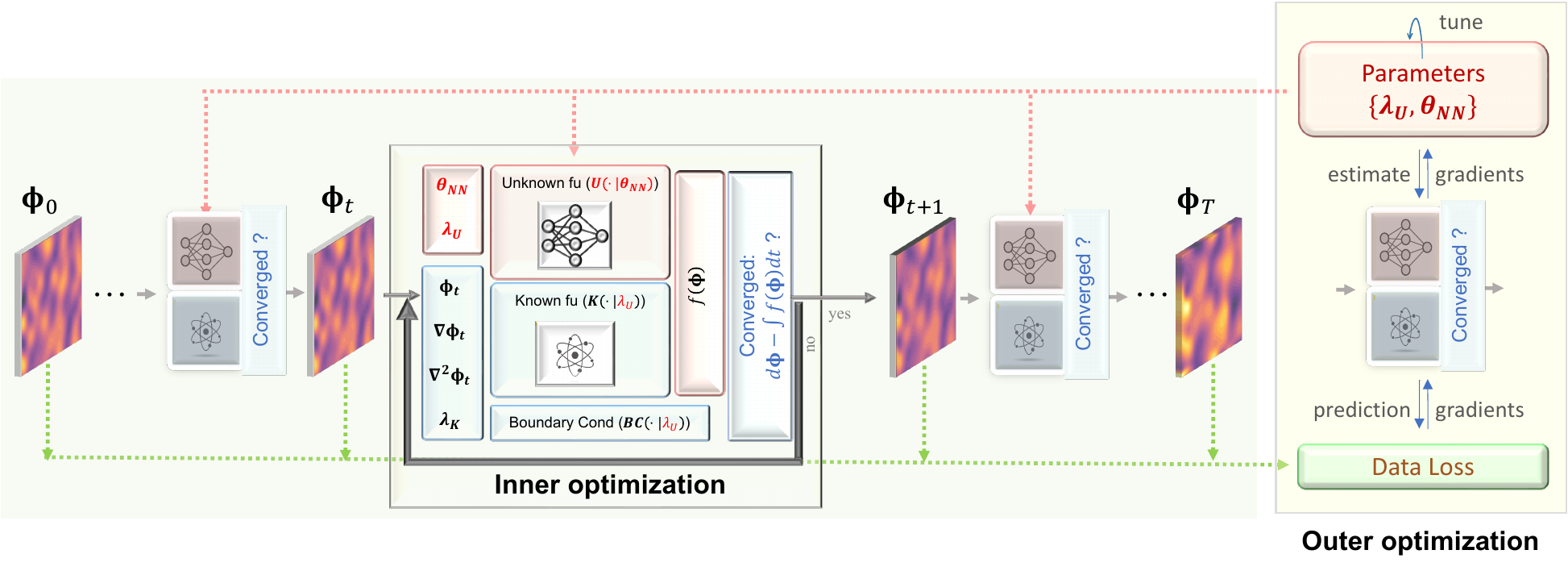}
    \caption{Schematic of the Im-PiNDiff framework, where temporal states $\boldsymbol{\Phi}_t$ are advanced via implicit updates incorporating known and learned physics. }
    \label{fig:DiffHybrid}
\end{figure}

To address these limitations, we introduce an implicit PiNDiff model, termed \textit{Im-PiNDiff}, inspired by implicit numerical schemes known for their superior numerical stability and robustness. Specifically, we replace explicit recurrent layers with an implicit layer formulation, analogous to the deep equilibrium model (DEQ) introduced by Bai et al.~\cite{bai2019deep}. DEQ employs implicit neural computations through solving a fixed-point equilibrium, enabling effectively infinite-depth representations without explicit unrolling of iterative layers. Similarly, in the context of hybrid neural-physics modeling, Im-PiNDiff conceptualizes each time step as solving an implicit nonlinear equation rather than explicit forward stepping. Training the Im-PiNDiff model involves explicitly embedding PDE constraints, resulting in a PDE-constrained bilevel optimization problem formulated mathematically as follows: 
\begin{subequations}\label{eq:Hybrid_opt}
\begin{align}
&\min_{\boldsymbol{\theta}} \quad J(\phi_{\boldsymbol{\theta}}(\mathbf{x}, t), \mathcal{D}; \boldsymbol{\theta}), \Omega_{p,t}, \\
\textrm{s.t.} \
& 
     \phi_{\boldsymbol{\theta}}(\mathbf{x},t_{i+1}) =  \phi_{\boldsymbol{\theta}}(\mathbf{x},t_{i}) + \int_{t_i}^{t_{i+1}} \mathscr{F}_{nn}\big[\phi_{\boldsymbol{\theta}}(\mathbf{x}, t); \boldsymbol{\theta}\big] dt, \quad i = 0, \dots, T \\
&
    \phi_{\boldsymbol{\theta}}(\mathbf{x}_{BC}, t_i) = \mathcal{BC}(\phi_{\boldsymbol{\theta}}(\mathbf{x}, t_i); \boldsymbol{\theta}), \quad i = 0, \dots, T\\
&
    \phi_{\boldsymbol{\theta}}(\mathbf{x}, t_0) = \mathcal{IC}(\boldsymbol{\theta}),\ \  \ \mathbf{x} \in \Omega_{p},
\end{align}
\label{eq:Hybrid_opt_Im}
\end{subequations}
where the outer-level optimization aims to minimize the discrepancy between model predictions and observed data (the loss function), while the inner-level optimization involves solving a nonlinear equilibrium equation to determine the solution $\phi_{\boldsymbol{\theta}}(\mathbf{x},t_{i+1})$ at each rollout step $t_{i+1}$. A schematics illustrating the Im-PiNDiff and its training strategy is presented in Fig. \ref{fig:DiffHybrid}
As shown in Fig.~\ref{fig:DiffHybrid}, propagating gradients through bilevel optimization poses significant challenges since standard AD frameworks typically require static computational graphs. Naively implementing unrolled differentiation with a fixed-iteration inner solver is computationally expensive and memory-intensive, severely limiting scalability. To circumvent this, we propose a hybrid training strategy integrating the adjoint-state implicit differentiation method with reverse-mode AD. Specifically, the adjoint-state approach propagates gradients through implicit equilibrium constraints, subsequently combined with reverse-mode AD gradients via the chain rule. A key benefit of adjoint-based gradient propagation is that it eliminates the necessity for storing intermediate computational nodes during forward propagation in implicit layers, significantly reducing memory usage. Additionally, we leverage checkpointing strategies to enhance memory efficiency during training. By combining adjoint methods with strategic checkpointing, our hybrid training strategy effectively balances computational performance and memory efficiency, enabling practical and scalable training of Im-PiNDiff models for large-scale scientific modeling problems.

\subsection{Adjoint-based hybrid AD for efficient gradient propagation}

Adjoint-based gradient computation methods have been increasingly adopted to enhance the efficiency and scalability of neural network training, especially for architectures involving implicitly defined operations~\cite{pan2023adjointdpm,matsubara2023symplectic,fidkowski2021adjoint,chen2018neural,bai2019deep}. In this study, we present an adjoint-based implicit differentiation approach for efficient Im-PiNDiff training, which require solving a nonlinear inner optimization during forward propagation.

\begin{figure}[!htp]
    \centering
    \includegraphics[width=0.90\linewidth]{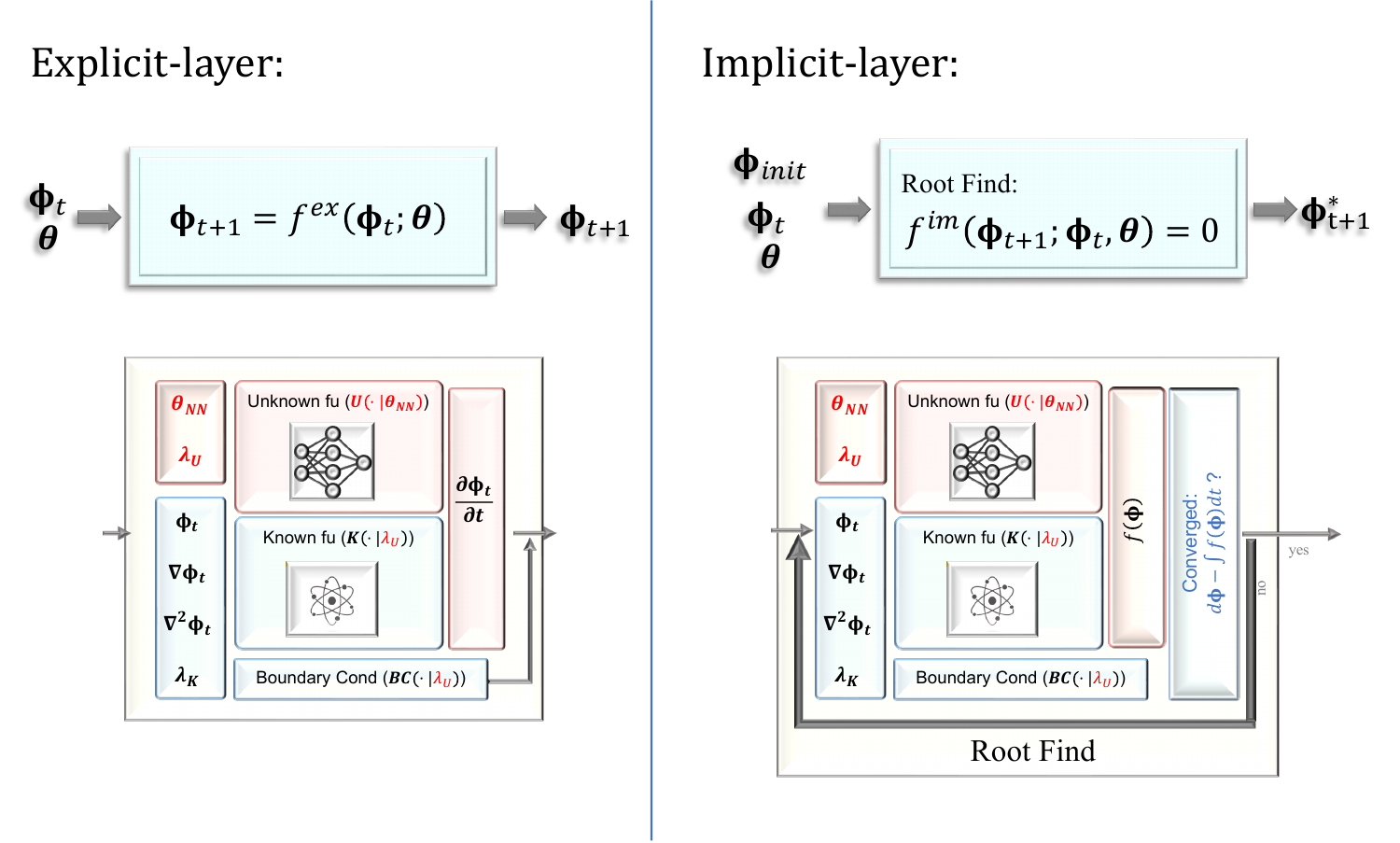}
    \caption{Comparison between the schematics of PiNDiff and Im-PiNDiff layers. }
    \label{fig:Im-layer}
\end{figure}

The fundamental distinction between explicit and implicit layers in PiNDiff models is illustrated in Fig.~\ref{fig:Im-layer}. 
We formally define an implicit layer by: 
\begin{equation}
    f^{im}\big(\boldsymbol{\Phi}_{t}; \boldsymbol{\Phi}_{t-1}, \boldsymbol{\theta}\big) = 0,
    \label{eq:implicit}
\end{equation}
where $\boldsymbol{\Phi}_{t} \in \mathbb{R}^n$ is the discrete state vector at the current step $t$ and  $\boldsymbol{\Phi}_{t-1}$ represents known previous states, dependent implicitly on parameter $\boldsymbol{\theta} \in \mathbb{R}^p$. The nonlinear implicit function $f^{im}: \mathbb{R}^n\times\mathbb{R}^n\times\mathbb{R}^p \to \mathbb{R}^n$. Given the typically high dimensionality of the problem, iterative numerical root-finding algorithms are usually employed to solve for $\boldsymbol{\Phi}_{t}$. Thus, the forward pass through the implicit layer can be succinctly expressed as,
\begin{equation}
    \boldsymbol{\Phi}_{t}^* \gets \text{RootFind}\Big( f^{im}\big(\boldsymbol{\Phi}_{t}; \boldsymbol{\Phi}_{t-1}, \boldsymbol{\theta}\big) = 0 \Big).
\end{equation}
DP with standard AD computes gradients by systematically applying the chain rule to decompose complex computer program into elementary operations. During the forward propagation, DP evaluates and records intermediate variables and their associated computational dependencies, building a computational graph from the inputs to the outputs. In reverse-mode AD, gradients are computed by traversing this computational graph backwards. Specifically, starting from the output, DP computes vector-Jacobian products (VJPs) at each node, sequentially applying the chain rule in reverse order. These VJPs represent the sensitivity of the output with respect to each intermediate variable. For the total loss $L(\boldsymbol{\theta})$ defined over the entire rollout trajectory of $T$ steps,
\begin{equation}
    L(\boldsymbol{\theta}) = \sum_{t=0}^{T}\mathcal{L}_t(\boldsymbol{\Phi}_t;\mathcal{D}_t) + \mathcal{L}_{regulate}(\boldsymbol{\theta}),
\end{equation}
the gradient of loss with respect to $\boldsymbol{\theta}$ can be computed using the chain rule,
\begin{equation}
    \frac{dL}{d\boldsymbol{\theta}} = \frac{\partial L}{
    \partial \boldsymbol{\theta}} + \sum_{t=0}^{T} \frac{\partial L}{
    \partial \boldsymbol{\theta} }\bigg|_{\boldsymbol{\Phi}_t^*}, 
\end{equation}
where
\begin{subequations}\label{eq:autoloss}
\begin{align}
    &\frac{\partial L}{
    \partial \boldsymbol{\theta} }\bigg|_{\boldsymbol{\Phi}_t^*} = \frac{\partial L}{ \partial \boldsymbol{\Phi}_{t+1}^*} \cdot \Bigg[\frac{\partial \boldsymbol{\Phi}_{t+1}^*}{ \partial \boldsymbol{\theta}} \Bigg],\\
    &\frac{\partial L}{\partial \boldsymbol{\Phi}_t^*} = \frac{\partial \mathcal{L}_t}{\partial \boldsymbol{\Phi}_{t}^*} + \frac{\partial L}{
    \partial \boldsymbol{\Phi}_{t+1}^*} \cdot \Bigg[\frac{\partial \boldsymbol{\Phi}_{t+1}^*}{ \partial \boldsymbol{\Phi}_{t}^*} \Bigg],
\end{align}
\end{subequations}
where $\frac{\partial L}{
    \partial \boldsymbol{\Phi}_{t+1}^*} \Big[\frac{\partial \boldsymbol{\Phi}_{t+1}^*}{
    \partial \boldsymbol{\theta}} \Big]$ and $\frac{\partial L}{
    \partial \boldsymbol{\Phi}_{t+1}^*} \Big[\frac{\partial \boldsymbol{\Phi}_{t+1}^*}{
    \partial \boldsymbol{\Phi}_{t}^*} \Big]$ are VJPs.
Given that the output $\boldsymbol{\Phi}^*_t$ from implicit layer is obtained by solving a nonlinear equation through an iterative numerical solver, the process can be abstractly written as a sequence of intermediate iterates:
\begin{equation}
    \Big \{\boldsymbol{\Phi}_{t}^{[0]}, \boldsymbol{\Phi}_{t}^{[1]}, \cdots, \boldsymbol{\Phi}_{t}^{[K]} \Big\} = \Big\{  \text{RootFind}^{[k]}\Big( f_{\boldsymbol{\theta}}^{im}\big(\boldsymbol{\Phi}_{t}^{[0]}; \boldsymbol{\Phi}_{t-1}\big) = 0 \Big) \Big\}_{k=0}^K
\end{equation}
where $K$ denotes the number of iterations needed to converge to the equilibrium solution $\boldsymbol{\Phi}^*_t \approx \boldsymbol{\Phi}_{t}^{[K]}, K \gg 1$. If one directly applies standard reverse-mode AD to propagate gradients through this implicit layer, the AD engine must record and retain the full computational graph of all intermediate iterates $\boldsymbol{\Phi}^{[k]}_t$ to compute VJPs. This results in substantial memory overhead, especially when the number of iterations or the state dimension $n$ is large.  

To address this limitation, we present a hybrid gradient computation strategy using the discrete adjoint-state method, which decouples gradient propagation from the forward iteration history. Instead of tracing all internal solver steps, the adjoint method efficiently computes the required VJP by solving a single linear system, avoiding the need to store intermediate iterates of the root-finding process. Concretely, for each time step $t \to t+1$, the Jacobians of $\boldsymbol{\Phi}_{t+1}^*$ with respect to $\boldsymbol{\Phi}_t^*$ and $\boldsymbol{\theta}$ are computed using the implicit function theorem~\cite{blondel2024elements},
\begin{subequations}\label{eq:implicitDiff}
\begin{align}
    &\frac{\partial \boldsymbol{\Phi}_{t+1}^*}{ \partial \boldsymbol{\theta}} = -\left(\frac{\partial f^{im}}{\partial \boldsymbol{\Phi}_{t+1}^*}\right)^{-1} \cdot \frac{\partial f^{im}}{\partial \boldsymbol{\theta}},\\
    &\frac{\partial \boldsymbol{\Phi}_{t+1}^*}{ \partial \boldsymbol{\Phi}_{t}^*} = -\left(\frac{\partial f^{im}}{\partial \boldsymbol{\Phi}_{t+1}^*}\right)^{-1} \cdot \frac{\partial f^{im}}{\partial \boldsymbol{\Phi}_{t}^*}.
\end{align}
\end{subequations}
Now we define the adjoint vector $\mathbf{w}^T_{t+1} \in \mathbb{R}^{1\times n}$ as,
\begin{equation}
    \mathbf{w}^T = -\frac{\partial L}{  \partial \boldsymbol{\Phi}_{t+1}^*} \cdot \left(\frac{\partial f^{im}}{  \partial \boldsymbol{\Phi}_{t+1}^*} \right)^{-1},
\end{equation}
and then the VJPs are expressed as,
\begin{subequations}\label{eq:implicitDiff}
\begin{align}
    &\frac{\partial L}{ \partial \boldsymbol{\Phi}_{t+1}^*} \cdot \Bigg[\frac{\partial \boldsymbol{\Phi}_{t+1}^*}{ \partial \boldsymbol{\theta}} \Bigg] = \mathbf{w}^T_{t+1} \cdot \frac{\partial f^{im}}{\partial \boldsymbol{\theta}},\\
    &\frac{\partial L}{ \partial \boldsymbol{\Phi}_{t+1}^*} \cdot \Bigg[\frac{\partial \boldsymbol{\Phi}_{t+1}^*}{ \partial \boldsymbol{\Phi}_{t}^*} \Bigg] = \mathbf{w}^T_{t+1} \cdot \frac{\partial f^{im}}{\partial \boldsymbol{\Phi}_{t}^*},
\end{align}
\end{subequations}
where $\mathbf{w}^T$ is obtained by solving the linear system,
    \begin{equation}
        \textbf{w}^T\bigg[\frac{\partial f^{im}}{  \partial \boldsymbol{\Phi}_{t+1}^*} \bigg]=-\frac{\partial L}{  \partial \boldsymbol{\Phi}_{t+1}^*}.
    \end{equation}
This linear equation is solved efficiently using iterative numerical linear solvers (such as GMRES or conjugate gradient methods). Instead of explicitly forming the Jacobian matrix $\frac{\partial f^{im}}{\partial \boldsymbol{\Phi}_t^*} \in \mathbb{R}^{n \times n}$ and $\frac{\partial f^{im}}{\partial\boldsymbol{\theta}} \in \mathbb{R}^{n \times p}$, we utilize VJP functions provided by AD tools to directly obtain $\mathbf{w}^T \frac{\partial f^{im}}{\partial \boldsymbol{\Phi}_t} \in \mathbb{R}^{1 \times n}$. Modern AD libraries (e.g., JAX, PyTorch, TensorFlow) directly provide VJP computations without forming the entire Jacobian matrix explicitly. Thus, the product $\mathbf{w}^T \frac{\partial f^{im}}{\partial \boldsymbol{\Phi}_t} \in \mathbb{R}^{1 \times n}$ is efficiently computed on-the-fly via AD-generated VJP functions, greatly enhancing computational and memory efficiency  (More details on the derivation of the hybrid adjoint backpropagation can be found in \ref{app:adj-VJP-derivation}).

The complete hybrid adjoint-based AD procedure for gradient back-propagation through an implicit layer can be summarized explicitly as follows: (1) solve the linear equation iteratively for $\mathbf{w}^T$; (2) compute the gradient with respect to parameters $\boldsymbol{\theta}$ and $\boldsymbol{\Phi}_{t}^*$ using the obtained adjoint vector and provide it to the AD to continue further backpropagation. 
\begin{figure}[!htp]
    \centering
    \includegraphics[width=\linewidth]{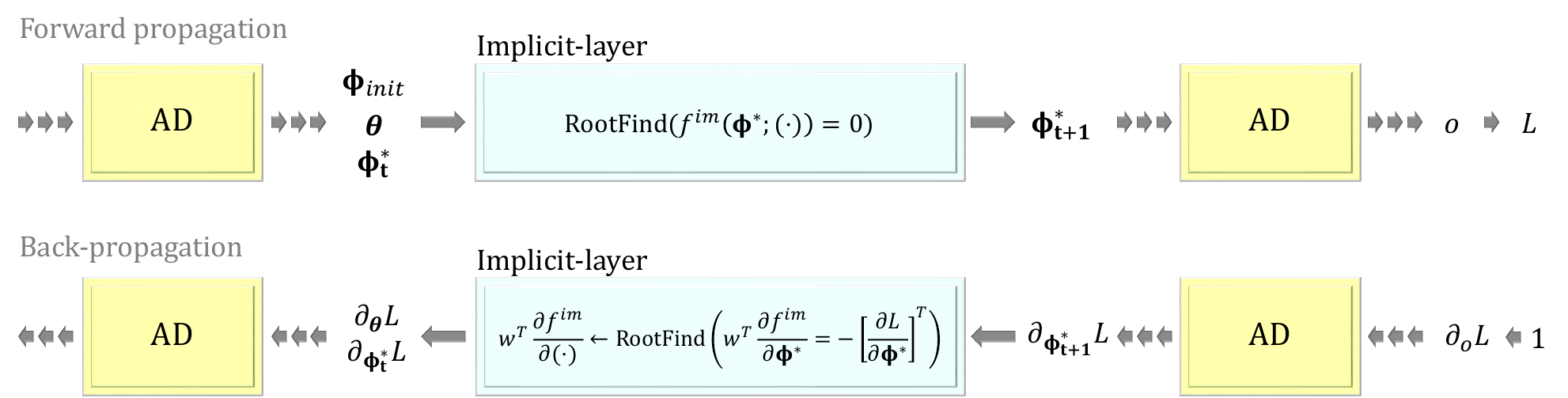}
    \caption{A schematic of forward and backward propagation through an implicit layer of the Im-PiNDiff model.}
    \label{fig:Im-layer-backprop}
\end{figure}
The schematic illustration of forward and backward passes through the implicit layer is provided in Fig.~\ref{fig:Im-layer-backprop}, and the complete algorithmic implementation steps are clearly outlined below in Algorithm~\ref{alg:Im-layer}.

\begin{algorithm}[H]
\footnotesize
\SetAlgoLined
\DontPrintSemicolon

    \SetKwFunction{FMain}{Forward}
    \SetKwProg{Fn}{Function}{:}{}
    \Fn{\FMain{$ \boldsymbol{\Phi}^*_{init}, \boldsymbol{\Phi}_t^*, \boldsymbol\theta$}}{

        $ \boldsymbol{\Phi}_{t+1}^* \gets \text{RootFind} \Big( f^{im}( \boldsymbol{\Phi}_{t+1};  \boldsymbol{\Phi}_t^*, \boldsymbol\theta) = 0 \Big)$
          
        \textbf{return} $\boldsymbol{\Phi}_{t+1}^*$ 
    }

\hspace{30 mm}


    \SetKwFunction{FMain}{Backward}
    \SetKwProg{Fn}{Function}{:}{}
    \Fn{\FMain{$\partial_{\boldsymbol{\Phi}_{t+1}^*}L, \boldsymbol{\Phi}_{t+1}^*, \boldsymbol{\Phi}_t^*, \boldsymbol\theta$}}{

        $\partial_{\boldsymbol{\Phi}_{t+1}^*} f^{im}(\cdot) \gets vjp(f^{im}(\cdot; \boldsymbol{\Phi}_t^*, \boldsymbol\theta))$ \Comment{$\partial_{b}a = \big[ \frac{\partial a}{\partial b}\big]$}
        
        $\partial_{\boldsymbol{\Phi}_{t}^*} f^{im}(\cdot) \gets vjp(f^{im}(\boldsymbol{\Phi}_{t+1}^*; \cdot, \boldsymbol\theta ))$ 
        
        $\partial_{\boldsymbol\theta} f^{im}(\cdot) \gets vjp(f^{im}(\boldsymbol{\Phi}_{t+1}^*; \boldsymbol{\Phi}_{t}^*,\cdot ))$ 
        
        $\textbf{w}^T_{t+1} \gets \text{RootFind} \Big(\partial_{\boldsymbol{\Phi}_{t+1}^*}f^{im}(\textbf{w}^T)  = - \partial_{\boldsymbol{\Phi}_{t+1}^*}L \Big)$ \Comment{$\partial f^{im}_{(\cdot)}(\textbf{w}^T) = \textbf{w}^T\big[ \frac{\partial f^{im}}{\partial (\cdot)}\big]$}
        
        $\partial_{\boldsymbol{\Phi}_{t+1}^*} L \cdot [\partial_{\boldsymbol{\Phi}_t^*} {\boldsymbol{\Phi}_{t+1}^*}] \gets \partial_{\boldsymbol{\Phi}_t^*} f^{im}(\textbf{w}^T_{t+1})$ 
        
        $\partial_{\boldsymbol{\Phi}_{t+1}^*} L \cdot [\partial_{\boldsymbol\theta} {\boldsymbol{\Phi}_{t+1}^*}] \gets \partial_{\boldsymbol\theta} f^{im}(\textbf{w}^T_{t+1})$ 

        \textbf{return} $ \partial_{\boldsymbol{\Phi}_{t+1}^*} L \cdot [\partial_{\boldsymbol{\Phi}_t^*} {\boldsymbol{\Phi}_{t+1}^*}], \ \ \partial_{\boldsymbol{\Phi}_{t+1}^*} L \cdot [\partial_{\boldsymbol\theta} {\boldsymbol{\Phi}_{t+1}^*}] $ 
    }
    
\caption{\small Algorithm for Implicit layer with adjoint-based backpropgation}
\label{alg:Im-layer}
\end{algorithm}

\subsection{Conditional neural fields for latent physics inference}
To capture spatially and temporally varying latent physical quantities, such as unresolved PDE terms, parametric fields, or unmodeled operators, we incorporate conditional neural fields (CNFs) into the Im-PiNDiff framework. Neural fields (NF), also known as coordinate-based implicit neural representations, offer a flexible and expressive mechanism for modeling continuous functions and operator-valued mappings over space and time. These representations have gained significant traction in computer vision, graphics, and scientific machine learning due to their ability to encode high-frequency and nonstationary features in data with minimal inductive bias~\cite{xie2022neural,sitzmann2020implicit,du2024conditional}. 

\begin{figure}[!htp]
    \centering
    \includegraphics[width=0.5\linewidth]{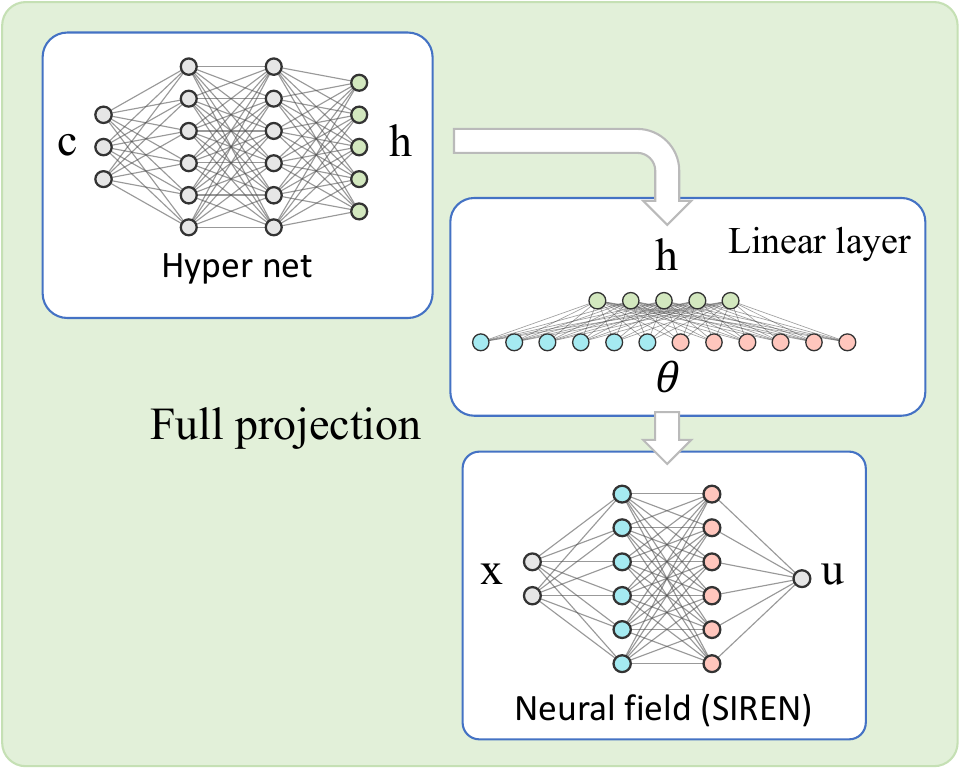}
    \caption{Architecture of the conditional neural field (CNF) module for latent physics inference. A hypernetwork maps a contextual input vector $\mathbf{c}$ to a latent embedding $\mathbf{h}$, which is linearly projected to generate the weights $\boldsymbol{\theta}_b$ of a SIREN-based neural field. }
    \label{fig:CNF}
\end{figure}

Within the context of PDE-constrained modeling, neural fields provide an elegant tool for parameterizing unknown coefficients or operators that vary over the spatial or spatiotemporal domain. The coordinate-continuous nature makes them particularly well suited for this task, as they can be queried at arbitrary spatial or temporal resolutions, ensuring mesh-invariant predictions and generalization to unseen domains. In this work, we employ a conditional formulation of neural fields, wherein the predicted physical field, e.g., unknown advection velocity $\mathbf{u}(\mathbf{x}, t)$ or diffusivity field $k(\mathbf{x})$, is conditioned on a latent embedding derived from auxiliary input. This conditioning allows the model to encode global contextual information, such as initial or boundary conditions, simulation settings, or observed response trajectories. As illustrated in Fig.~\ref{fig:CNF}, our architecture comprises three modules: a hypernetwork that maps condition vectors to latent codes, a linear projector that translates latent codes to NF parameters, and a base NF network that evaluates the inferred field at queried coordinates.

Specifically, given a conditioning vector $\mathbf{c}$ representing contextual input (e.g., initial/boundary condition encodings or low-dimensional representations of observed dynamics), a hypernetwork $\text{HyperNet}_{\boldsymbol{\theta}_h}$ produces a latent code $\mathbf{h} \in \mathbb{R}^{d_L}$, 
\begin{equation} 
    \mathbf{h} = \text{HyperNet}_{\boldsymbol{\theta}_h}(\mathbf{c}), 
\end{equation}
which is subsequently projected via a learnable linear operator into the NF parameters $\boldsymbol{\theta}_{b} = \mathbf{W}_{\text{proj}}\cdot \mathbf{h}$. The base NF network, implemented as a sinusoidal representation network (SIREN)~\cite{sitzmann2020implicit}, then evaluates the spatial or spatiotemporal field as 
\begin{equation} 
\mathbf{u}(\mathbf{x}; \mathbf{c}) = \text{SIREN}_{\boldsymbol{\theta}_{b}}(\mathbf{x}), 
\end{equation} 
where $\mathbf{u}(\cdot)$ represents the inferred latent field, such as a velocity or forcing term. The use of SIREN allows the model to resolve fine-scale variations in the latent field and capture complex physical patterns across domains with smooth and expressive function approximations.

Crucially, the conditional neural field $\mathbf{u}_{\boldsymbol{\theta}_{b}}(\mathbf{x}; \mathbf{c})$ is mixed directly with the known PDE operators to form the hybrid neural PDE operator $\mathscr{F}_{nn}[\cdot]$ in the Im-PiNDiff solver, allowing it to affect the system dynamics during training. Because the CNF module is fully differentiable and compatible with our adjoint-based gradient propagation strategy, end-to-end learning remains tractable and memory-efficient. This integration enables the identification of latent physics from sparse indirect observations. That is, only a small part of the state trajectories $\{\boldsymbol{\Phi}_t\}_{t=0}^{T}$ are observable, and the physical field itself is hidden.

\section{Results and Discussion}
\label{sec:results}

\subsection{Forward and nnverse modeling of spatiotemporal physics}
We assessed the proposed Im-PiNDiff framework on two canonical spatiotemporal PDE systems: the advection-diffusion equation and the scalar Burgers' equation. These case studies demonstrate the model's capability to perform both forward prediction and inverse inference of latent physical fields under linear and nonlinear dynamics, respectively. In particular, the advection–diffusion system serves as a testbed for investigating the accuracy and stability of Im-PiNDiff in handling smooth transport-diffusion processes, while the Burgers' equation probes its robustness in capturing nonlinear wave propagation and shock formation.

\subsubsection{Advection–diffusion processes with steady advection fields}
We begin with a 2D advection–diffusion system, which describes the spatiotemporal evolution of a scalar field $\phi(\mathbf{x}, t)$ under combined effects of directional transport and diffusion,
\begin{equation}
    \frac{\partial \phi}{\partial t} = -{u}_x \frac{\partial \phi}{\partial x} - {u}_y \frac{\partial \phi}{\partial y} + k \frac{\partial^2 \phi}{\partial x^2} + k \frac{\partial^2 \phi}{\partial y^2},
    \label{eq:AD}
\end{equation}
where $k$ is the diffusion coefficient and ${u}_x$, ${u}_y$ are the advection coefficients in the $x$ and $y$ directions, respectively. Here, ${u}_x(\mathbf{x})$ and ${u}_y(\mathbf{x})$ are treated as unknown spatial fields to be inferred during Im-PiNDiff training, where partial observations of the state variable $\phi(\mathbf{x}, t)$ are used as labels. These unknown fields are parameterized by CNFs.

To generate training and testing data, we employed a high resolution finite-volume (FV) solver with fourth-order Runge–Kutta (RK4) time integration and a time step of $10^{-3}$ s. The physical domain of size $2 \times 1$ was discretized using a $128 \times 64$ Cartesian mesh. A set of ten initial conditions was randomly generated from Gaussian processes (GPs) on a coarse grid $30 \times 10$ and projecting them onto the simulation mesh. These GP fields were constructed using radial basis kernels with varying length scales to promote diversity in initial configurations. Ground truth advection velocity fields were generated as smooth superpositions of cosine functions (details in \ref{sec:sinfu}) and similarly projected to the simulation grid. For training, only a very limited number of state observations, specifically, two snapshots of $\phi$, were provided, simulating a data-sparse regime. The model was then evaluated on unseen initial conditions to assess its ability to reconstruct the full spatiotemporal state ($\phi$) trajectories over extended prediction horizons, while simultaneously inferring the hidden advection velocity fields from these sparse, indirect observations.   

The Im-PiNDiff model was constructed using the FV discretization of the governing PDEs, where the unknown spatial advection velocity fields are modeled as CNFs parameterized continuously over space and conditioned on time. To achieve stable and accurate long-term state propagation, the temporal autoregression is through the implicit layer using a second-order Crank–Nicolson scheme,
\begin{equation}
    \boldsymbol\Phi_{t_{i+1}}(\boldsymbol{\theta}) = \boldsymbol\Phi_{t_i}(\boldsymbol{\theta}) + \frac{t_{i+1}-t_{i}}{2} \Big[ \mathscr{F}_{nn}\big[\boldsymbol\Phi_{t_{i+1}}(\boldsymbol{\theta}); \boldsymbol{\theta}\big] + \mathscr{F}_{nn}\big[\boldsymbol\Phi_{t_{i}}(\boldsymbol{\theta}); \boldsymbol{\theta}\big]\Big], \quad i = 0, \dots, T,
    \label{Eq:Crank-Nicolson}
\end{equation}
propagating state from $t_{i}$ to $t_{i+1}$. The neural model uses a relative large time step of $10^{-2}$ s, an order of magnitude larger than the step size used in generating the training data. At each time step, Eq.~\eqref{Eq:Crank-Nicolson} requires solving a nonlinear system due to its implicit dependence on $\boldsymbol\Phi_{t_{i+1}}$, which is accomplished via the Biconjugate Gradient Stabilized (BiCGStab) method. To enable efficient end-to-end training, we applied the hybrid adjoint-based AD backpropagation algorithm introduced in Section~\ref{sec:methodology}. 

\begin{figure}[!htp]
    \centering
    \includegraphics[width=\linewidth]{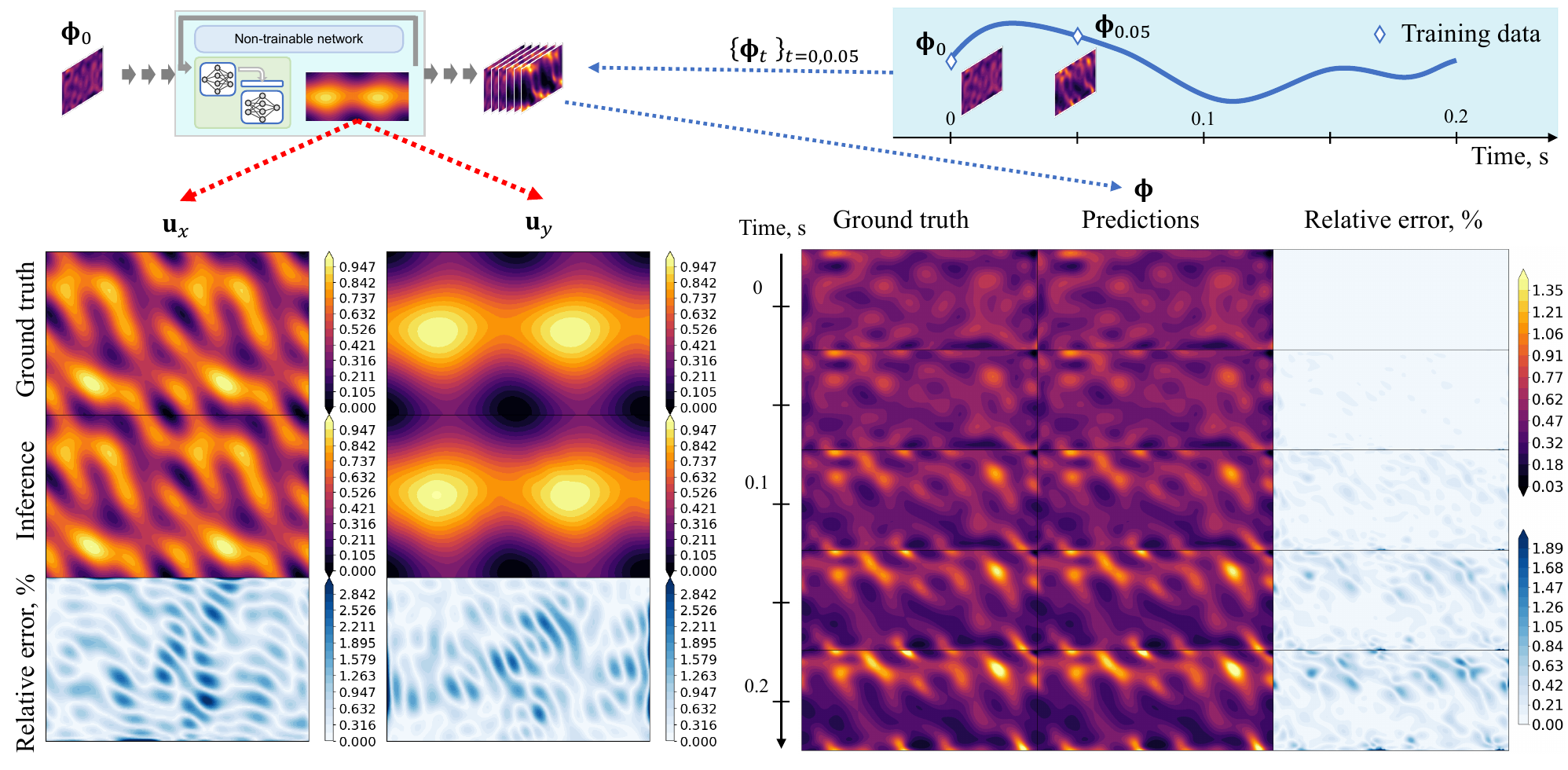}
    \caption{Forward prediction and inverse inference using Im-PiNDiff on the advection–diffusion problem with steady advection fields. The model is trained using only two snapshots ${\boldsymbol{\phi}t}{t=0, 0.05}$. Left: inferred steady advection fields $u_x(\mathbf{x})$ and $u_y(\mathbf{x})$ compared with ground truth. Right: predicted scalar field $\phi(\mathbf{x}, t)$ at future times, showing excellent agreement with ground truth across the forecast horizon. }
    \label{fig:AdDfStd}
\end{figure}

We first evaluated the Im-PiNDiff framework for the scenario where the advection velocity fields were assumed to be spatially varying but temporally invariant, i.e., ${u}_x = {u}_x(\mathbf{x})$ and ${u}_y = {u}_y(\mathbf{x})$. The model was trained using only two snapshots of the state field: the initial condition $\boldsymbol\Phi_0$ and an observation at $t = 0.05$ s, as indicated in the top-right panel of Fig.~\ref{fig:AdDfStd}.
After 2000 epochs of training, the model produced accurate forward predictions of the scalar field $\boldsymbol{\phi}(\mathbf{x}, t)$ over the extended horizon ${0.05, 0.1, 0.15, 0.2}$ s and simultaneously inferred the underlying steady advection fields. As shown in Fig.~\ref{fig:AdDfStd}, the predicted scalar field trajectories closely match the ground truth with a relative error of approximately 2\%, while the advection fields ${u}_x$ and ${u}_y$ are accurately inferred as well, with 3\% error compared to the true fields.

\begin{figure}[!htp]
    \centering
    \includegraphics[width=\linewidth]{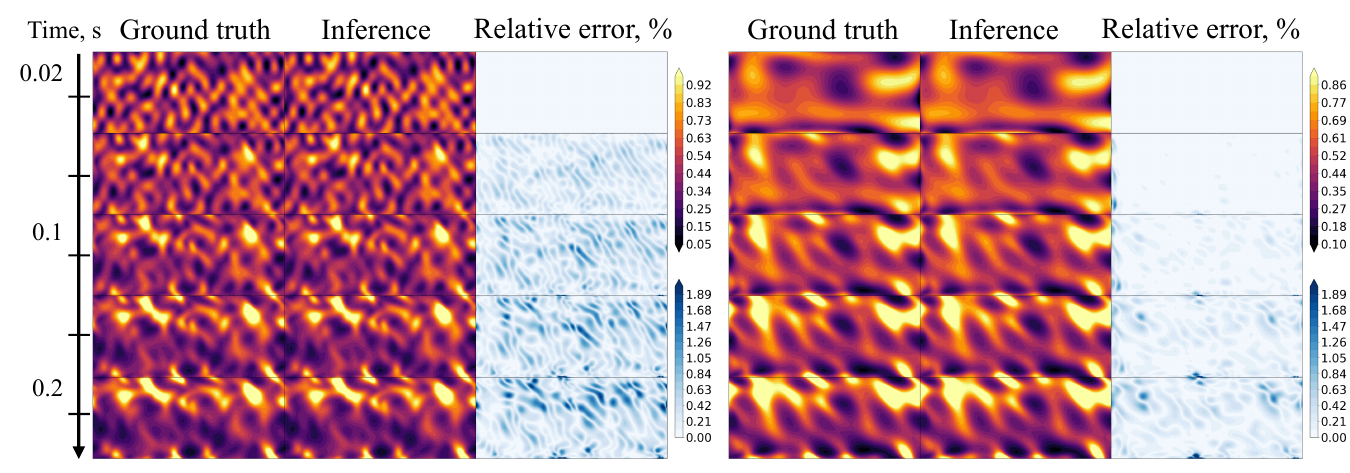}
    \caption{Im-PiNDiff predictions of $\phi(\mathbf{x}, t)$ for testing initial conditions with time-invariant advection fields.}
    \label{fig:AdDfStd_test}
\end{figure}
The trained model was further tested on out-of-distribution (OOD) initial conditions generated using radial basis kernels with characteristic length scales different from those used during training. As illustrated in Fig.~\ref{fig:AdDfStd_test}, the model accurately predicted the spatiotemporal evolution of the scalar field $\boldsymbol{\phi}(\mathbf{x}, t)$ over the full time horizon $[0, 0.2]$ s, despite having never seen such initial states during training. The predictions remain stable and accurate across time, with relative errors generally below 2\%, indicating strong robustness to randomly generated unseen initial conditions.

These results highlight the model's capacity to perform robust forward/inverse modeling from extremely limited data, accurately reconstructing both state dynamics and hidden physics. Notably, no direct observations of the advection fields were used during training, underscoring the effectiveness of the CNF and hybrid adjoint-AD gradient propagation over the entire program in inferring latent physics from sparse indirect measurements.

\subsubsection{Advection–diffusion processes with dynamic advection fields}

To further evaluate the robustness of Im-PiNDiff in modeling non-stationary physics, we consider a more challenging setting where the advection fields vary in both space and time, i.e., ${u}_x(\mathbf{x}, t)$ and ${u}_y(\mathbf{x}, t)$. This setting poses a significant challenge for spatiotemporal inversion, as the time-varying latent fields are inferred from sparse and irregular data.

\begin{figure}[!htp]
    \centering
    \includegraphics[width=\linewidth]{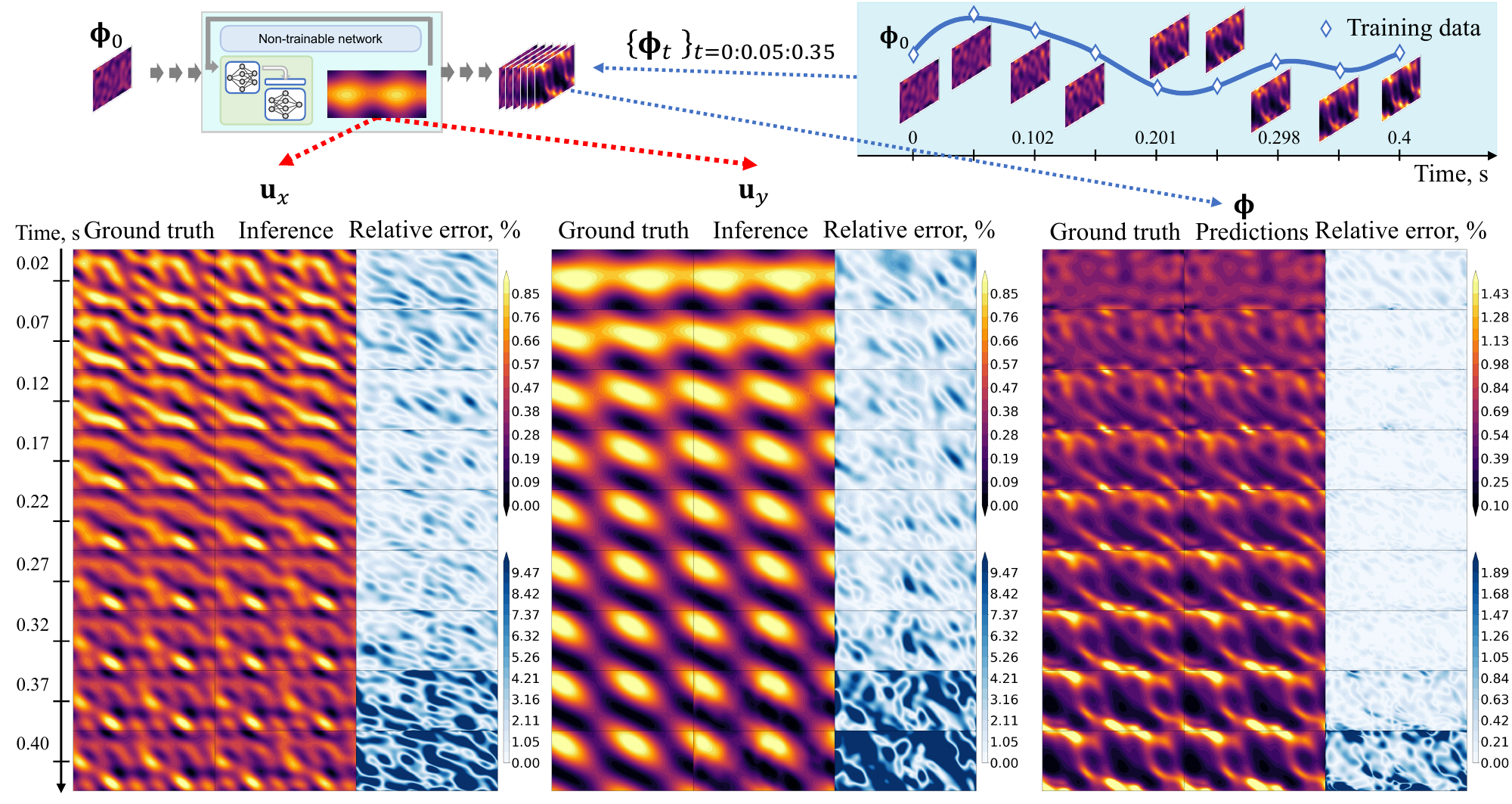}
    \caption{Forward prediction and inverse inference using Im-PiNDiff on the advection–diffusion problem with time-varying advection fields. }
    \label{fig:AdDfDyn}
\end{figure}
\begin{figure}[!htp]
    \centering
    \includegraphics[width=\linewidth]{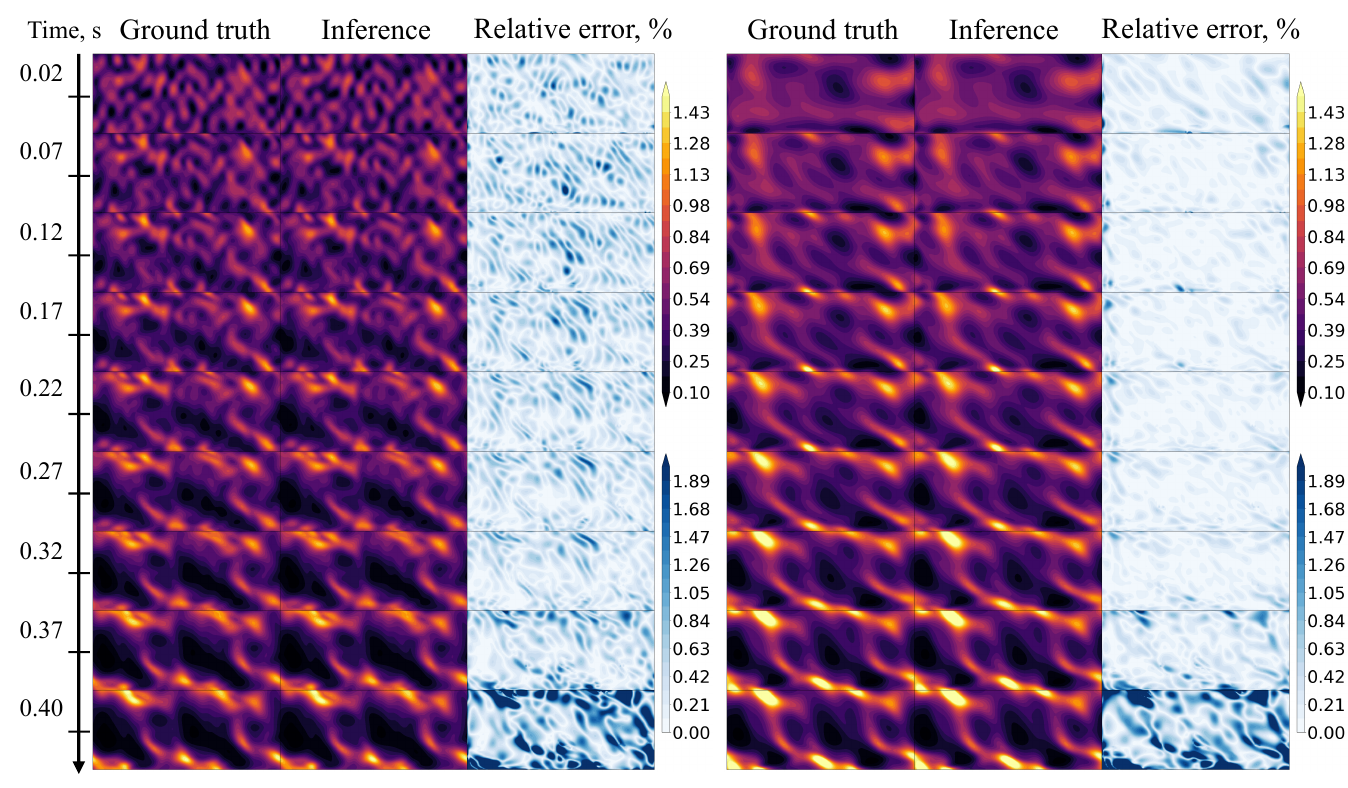}
    \caption{Im-PiNDiff predictions of $\phi(\mathbf{x}, t)$ for testing initial conditions with time-varying advection fields.}
    \label{fig:AdDfDyn_test}
\end{figure}

The training set consists of eight snapshots sampled at nonuniform time intervals ${t}_{\text{train}} = {0, 0.05, 0.102, 0.15, 0.201, 0.25, 0.298, 0.35}$ s over a total simulation horizon of $T = 0.4$ s. This choice reflects realistic scenarios where sensor data may be irregularly sampled in time. The model is trained for 10,000 epochs to jointly learn both the forward state evolution $\phi(\mathbf{x}, t)$ and the hidden time-varying advection fields ${u}_x(\mathbf{x}, t)$ and ${u}_y(\mathbf{x}, t)$. The learned model is then used to reconstruct full spatiotemporal fields at unobserved time points $t = {0.02, 0.07, 0.12, 0.17, 0.22, 0.27, 0.32, 0.37, 0.4}$, which are also sampled irregularly.

As shown in Fig.~\ref{fig:AdDfDyn}, the Im-PiNDiff model accurately recovers the ground truth dynamics and latent velocity fields. The left panels show the inferred ${u}_x$ and ${u}_y$ fields in comparison with their ground truth counterparts, along with relative error maps across time. Despite the inherent ill-posedness of recovering time-dependent vector fields from limited state observations, the inferred advection fields capture the spatial-temporal structures reasonably well, with a relative error around 10\%. The right panels show the predicted scalar field $\phi(\mathbf{x}, t)$, achieving high accuracy with a relative error consistently below 2\% across all prediction times. Generalization performance is further evaluated using OOD initial conditions, generated from unseen Gaussian process realizations. The predictions on these OOD cases, shown in Fig.~\ref{fig:AdDfDyn_test}, demonstrate excellent agreement with ground truth, confirming that the Im-PiNDiff model generalizes robustly to new initializations with high accuracy.

\subsubsection{Scalar Burgers' dynamics with spatially varying viscosity fields}

We further evaluate the Im-PiNDiff framework on the 2D scalar Burgers’ equation to test its capability in recovering spatially varying latent physical parameters under nonlinear dynamics,
\begin{equation}
    \frac{\partial {u}}{\partial t} = -{u} \frac{\partial {u}}{\partial x} - {u} \frac{\partial {u}}{\partial y} + {\nu} \frac{\partial^2 {u}}{\partial x^2} + {\nu} \frac{\partial^2 {u}}{\partial y^2},
\end{equation}
where ${u}(\mathbf{x}, t)$ is the scalar velocity field and ${\nu}(\mathbf{x})$ denotes the unknown spatially varying viscosity field. In this setting, the convective nonlinearity leads to steep gradients and localized structures, making the recovery of latent viscosity fields from sparse observations particularly challenging. Note that the viscosity is set to be of order $10^{-2}$; higher values would suppress the convective dynamics and render the solution diffusion-dominated.

The Im-PiNDiff model is trained using only four snapshots of the velocity field, without any direct supervision on the viscosity. The CNF module is tasked with recovering $\boldsymbol{\nu}(\mathbf{x})$ from the observed dynamics. 
\begin{figure}[!t]
    \centering
    \includegraphics[width=\linewidth]{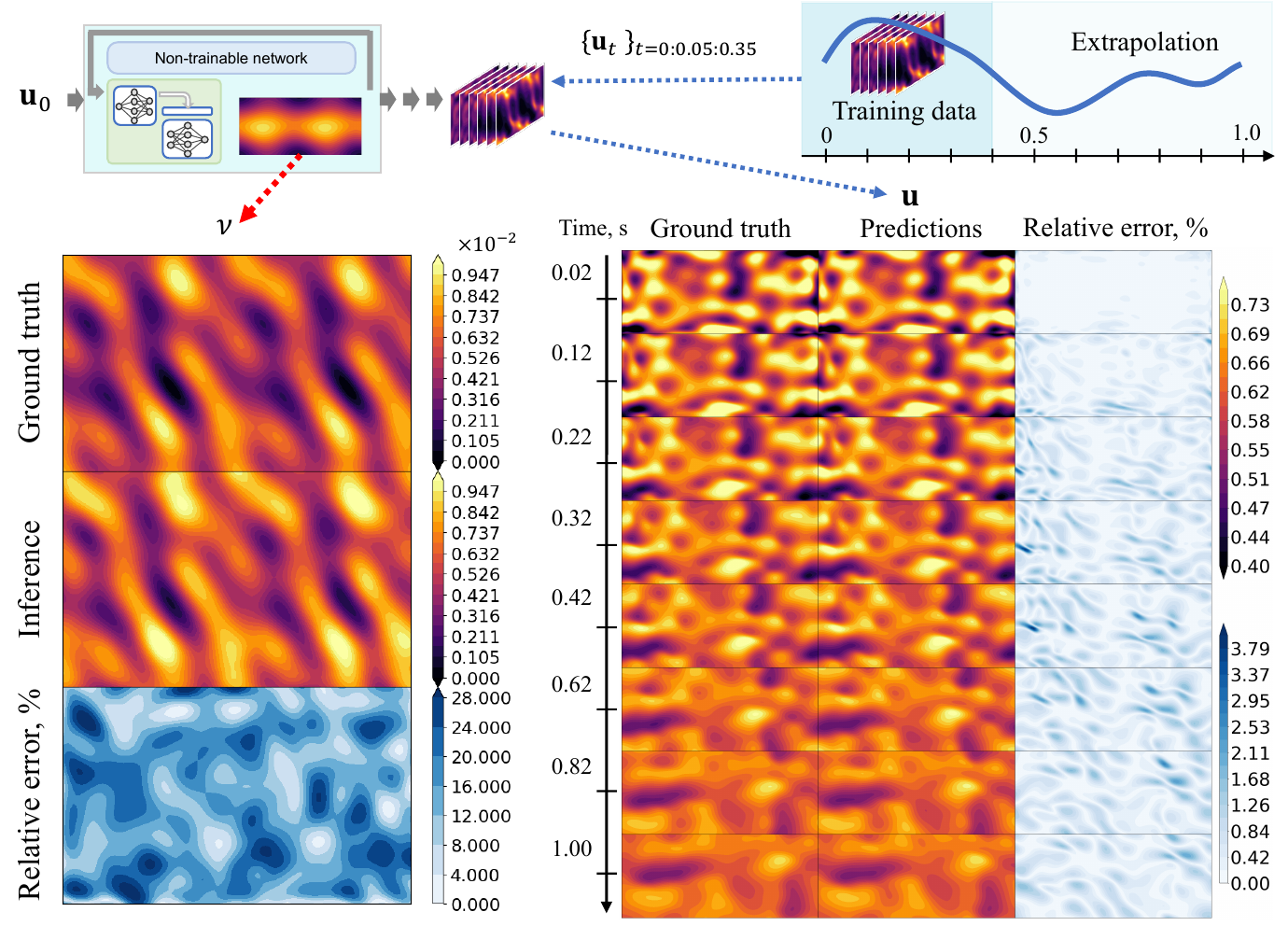}
    \caption{Inference of spatially varying viscosity field $\nu(\mathbf{x})$ and prediction of velocity field ${u}(\mathbf{x}, t)$ for the 2D scalar Burgers' equation using Im-PiNDiff. The model is trained on sparse observations of ${u}$ and successfully reconstructs both the hidden viscosity field and the spatiotemporal dynamics.}
    \label{fig:BurgStd}
\end{figure}
Figure~\ref{fig:BurgStd} shows the results of Im-PiNDiff for the scalar Burgers' dynamics with spatially varying viscosity. The left panel presents the inferred steady viscosity field $\nu(\mathbf{x})$ in comparison with the ground truth, alongside its corresponding relative error. Although recovering viscosity from indirect state observations is highly ill-posed, particularly in this regime where $\nu(\mathbf{x})$ is of order $10^{-2}$ and exerts only a weak influence on the dynamics, the inferred $\boldsymbol{\nu}(\mathbf{x})$ captures the dominant spatial patterns and exhibits reasonable structural agreement with the true field.  Some deviation is observed in regions of low sensitivity, resulting in higher relative errors (up to 28\%). The low sensitivity is evident from the velocity prediction results. As shown in the right panel, the model achieves highly accurate predictions of the velocity field, with a relative error around 4\%. These results demonstrate that Im-PiNDiff remains robust in recovering hidden spatial parameters while preserving predictive fidelity in nonlinear PDE systems. 

Further, the trained model was tested on unseen initial conditions.
\begin{figure}[t!]
    \centering
    \includegraphics[width=\linewidth]{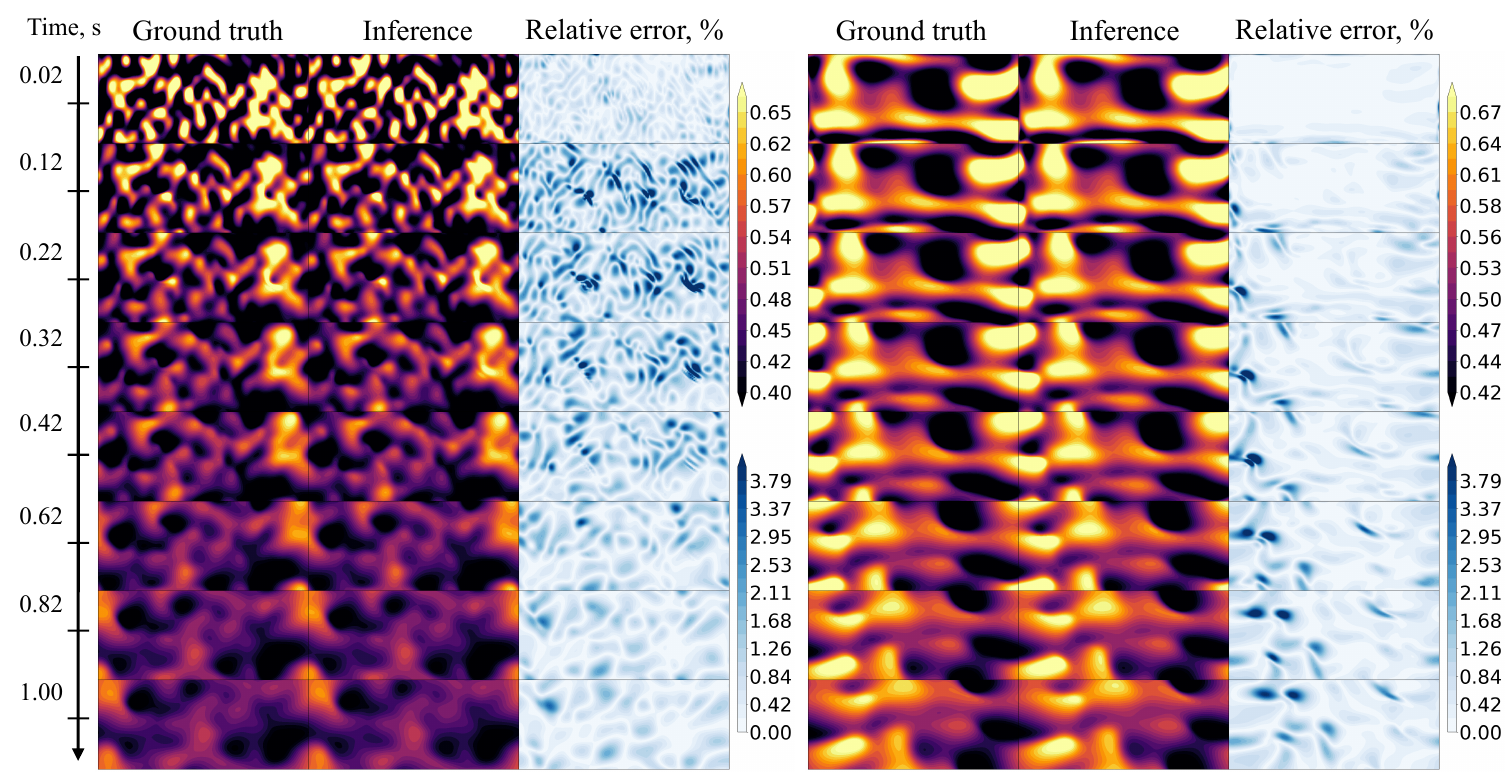}
    \caption{Im-PiNDiff predictions of Burgers' dyanmics on two testing (unseen) initial conditions.}
    \label{fig:BurgStd_test}
\end{figure}
 As shown in Fig.~\ref{fig:BurgStd_test}, the model accurately predicts the spatiotemporal evolution of the velocity field $\mathbf{u}(\mathbf{x}, t)$ across both test cases, achieving relative errors consistently around 4\%. These results confirm that the proposed framework generalizes robustly to new initializations, despite being trained under limited data and an ill-posed inference setting.

\subsection{Temporal error accumulation and stability analysis}

\begin{figure}[!t]
    \centering
    \subfloat[\centering Accumulation of relative error in the $\Phi_t$ over simulation time]{{\includegraphics[width=0.5\linewidth]{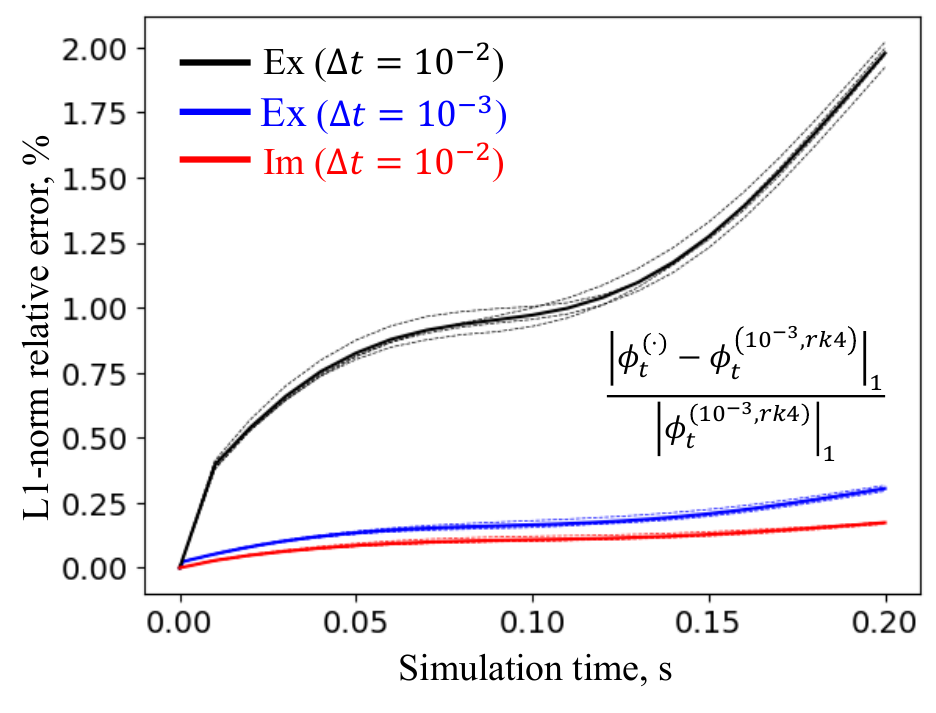} }}%
    \subfloat[\centering L1 relative error in the $\Phi_{0.2}$ at last time for various time-step sizes]{{\includegraphics[width=0.5\linewidth]{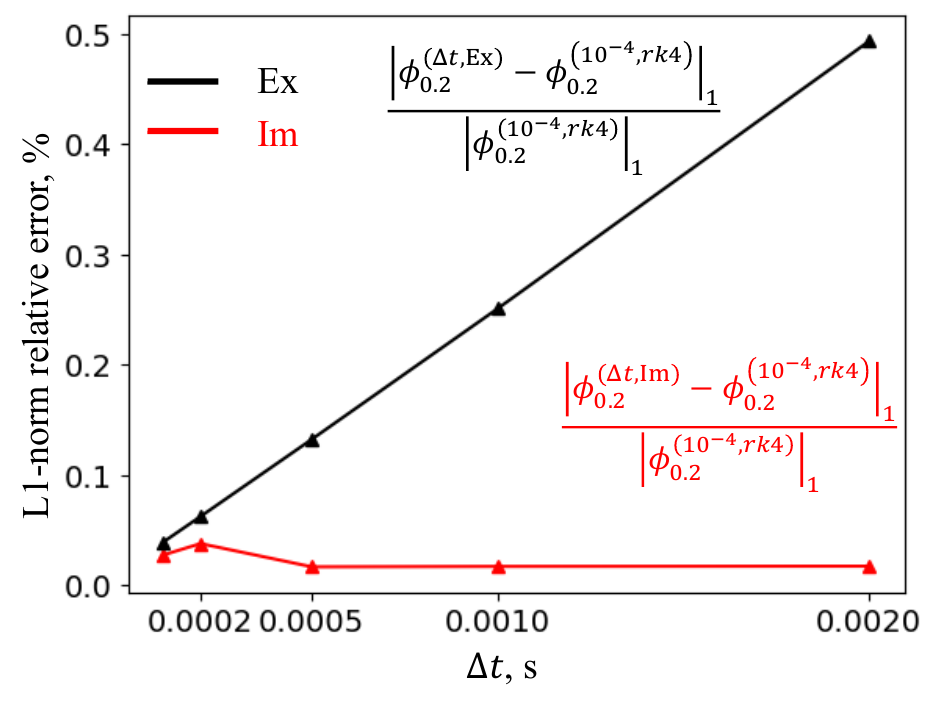} }}%
    \caption{Accumulation of relative error in the $\phi$ fields for the Ex-PiNDiff (Ex) and Im-PiNDiff (Im) models, evaluated on various test cases of Advection-Diffusion.}
    \label{fig:ErrAcc}%
\end{figure}

An important motivation for the proposed Im-PiNDiff framework lies in its ability to perform robust learning and stable long-horizon forecasting, while mitigating error accumulation. This advantage stems from the use of implicit autoregressive forward passes, which have long been favored in classical numerical analysis for their superior stability properties, especially in stiff or convection-dominated systems. To systematically quantify and compare the temporal error accumulation behavior, we conduct a controlled study using the advection–diffusion problem with steady advection fields. We benchmark the performance of Im-PiNDiff against its explicit counterpart, referred to as Ex-PiNDiff, which follows the original PiNDiff formulation with a recurrent explicit forward pass~\cite{akhare2023physics}. Both models are trained on the same dataset using the same model architecture, differing only in the time-stepping mechanism. We investigate two key aspects: (1) error accumulation over time for a fixed time-step size and (2) sensitivity of the final prediction error to varying temporal resolutions. The results are shown in Fig.~\ref{fig:ErrAcc}. 

Figure~\ref{fig:ErrAcc}(a) reports the accumulation of L1-norm relative error in the scalar field $\phi$ over the simulation horizon of $T = 0.2$ s, comparing three models: Ex-PiNDiff with $\Delta t = 10^{-2}$ s (black), Ex-PiNDiff with $\Delta t = 10^{-3}$ s (blue), and Im-PiNDiff with $\Delta t = 10^{-2}$ s (red). 
As the simulation progresses, the explicit model with coarse time steps ($\Delta t = 10^{-2}$ s) exhibits significant error growth, reflecting compounding integration and learning errors. In contrast, the implicit Im-PiNDiff maintains a nearly constant error profile throughout the rollout, highlighting its temporal stability. Notably, the implicit model achieves comparable or better accuracy than the explicit model even when trained at a tenfold coarser temporal resolution. 
Figure~\ref{fig:ErrAcc}(b) further quantifies the effect of time-step size on final prediction accuracy by plotting the L1 relative error in $\phi$ at $t = 0.2$ s across various $\Delta t$ values. The explicit model shows a steep increase in error as $\Delta t$ increases, underscoring its sensitivity to temporal resolution. In contrast, Im-PiNDiff remains remarkably robust, showing negligible degradation in accuracy even as $\Delta t$ grows. This result confirms that the implicit layer stabilizes long-horizon predictions, suppresses numerical drift, and allows for larger autoregressive steps without compromising predictive quality.

Taken together, these findings illustrate the key strength of the Im-PiNDiff framework in preserving predictive accuracy and stability over extended time horizons. By decoupling learning stability from time-step resolution, Im-PiNDiff not only ensures physically consistent forecasting but also enables significant computational savings, a benefit explored further in the subsequent section on computational cost analysis.

\subsection{Computational cost analysis}

To understand the efficiency and scalability of Im-PiNDiff models, we analyze the computational and memory complexity of Im-PiNDiff relative to Ex-PiNDiff under various settings.  In the Ex-PiNDiff model, the system evolves through explicit autoregressive updates, forming a forward trajectory of the form,
\begin{equation}
    \boldsymbol{\Phi}_{0} \rightarrow \boldsymbol{\Phi}_{\Delta t} \rightarrow \dots \rightarrow \boldsymbol{\Phi}_{t} \rightarrow \boldsymbol{\Phi}_{t +\Delta t} \rightarrow \dots \rightarrow \boldsymbol{\Phi}_{T -\Delta t} \rightarrow \boldsymbol{\Phi}_{T}\rightarrow L,
\end{equation}
where each transition step requires storing the full state $\boldsymbol{\Phi}_t \in \mathbb{R}^n$ and its associated computational graph for backpropagation through time. In contrast, the Im-PiNDiff model propagates the state implicitly by solving a nonlinear fixed-point equation at each time step. A naive AD-based implementation that unrolls $K$ solver iterations per step as,
\begin{equation}
    \boldsymbol{\Phi}_0 \rightarrow \big\{\boldsymbol{\Phi}_{\Delta t}^{[0]}  \cdots \boldsymbol{\Phi}_{\Delta t}^{[K]} \big\}\rightarrow\boldsymbol{\Phi}_{\Delta t}^* \cdots \big\{\boldsymbol{\Phi}_{t}^{[0]}  \cdots \boldsymbol{\Phi}_{t}^{[K]} \big\}\rightarrow\boldsymbol{\Phi}_{t}^* \cdots\big\{\boldsymbol{\Phi}_{T}^{[0]} \cdots \boldsymbol{\Phi}_{T}^{[K]} \big\}\rightarrow\boldsymbol{\Phi}_{T}^* \rightarrow L.
\end{equation}
where each $\boldsymbol{\Phi}{t}^{[k]}$ denotes the $k$-th iteration of the nonlinear solver at time $t$. This naive Im-PiNDiff approach incurs memory and runtime costs that scale linearly with both the number of time steps and the number of solver iterations $K$. To overcome this limitation, the proposed hybrid training strategy combines adjoint-state methods with reverse-mode AD, avoiding the need to store intermediate iterations by solving a single linear system via the implicit function theorem. The number of solver steps $\tilde{K}$ can be determined dynamically via convergence criteria. This approach decouples memory usage from $\tilde{K}$ while still retaining the flexibility of accurate, adaptive inner solvers, yielding significant gains in both scalability and efficiency. Table~\ref{tab:Comp_cost} summarizes the computational complexity of different strategies in terms of memory usage and training time.
\begin{table}[]
    \centering
    \small
    \begin{tabular}{c|c c}
        \hline
                   & Memory footprint & Training time \\
        \hline
        Ex-PiNDiff & $O\big((T/\Delta t) n\big)$ & $O\big((T/\Delta t)n^3\big)$ \\
        Naive Im-PiNDiff & $O\big((T/\Delta t)Kn\big)$ & $O\big((T/\Delta t)Kn^3\big)$ \\
        Im-PiNDiff & $O\big((T/\Delta t)n\big)$ & $O\big((T/\Delta t)\tilde{K}n^3\big)$ \\
        \hline
    \end{tabular}
    \caption{Computational cost for Ex-PiNDiff and Im-PiNDiff models.}
    \label{tab:Comp_cost}
\end{table}
The key observation is that the memory footprint of the hybrid Im-PiNDiff model is independent of $\tilde{K}$, 
\begin{wrapfigure}{r}{0.5\textwidth}
    \centering
    \vspace{-1.5em}
    \includegraphics[width=0.85\linewidth]{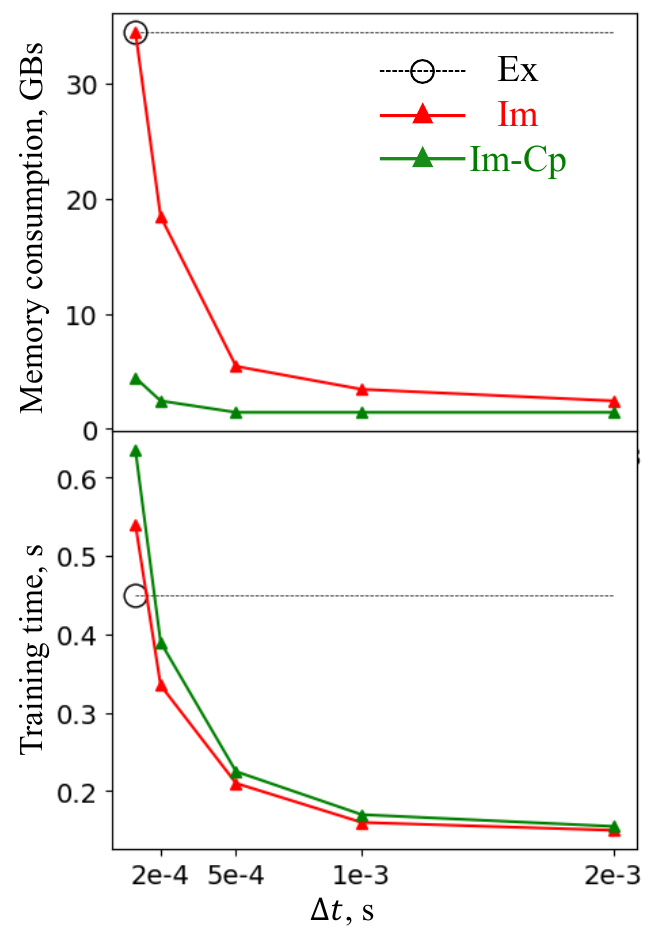}\vspace{-1em}
    \caption{Memory consumption and training time for the advection-diffusion case. 
    (\textcolor{black}{Ex}: Ex-PiNDiff, \textcolor{red}{Im}: Im-PiNDiff, \textcolor{green}{Im-Cp}: Im-PiNDiff w/ Checkpoint.)}\vspace{-1em}
    \label{fig:CompCost}
\end{wrapfigure}
unlike the naive version. This enables substantial memory savings without compromising expressivity or stability. Furthermore, since $\tilde{K}$ is adaptively determined, the total training time is often lower than that of the fixed-$K$ naive strategy.

Figure~\ref{fig:CompCost} presents the empirical computational cost for the advection–diffusion case, showing peak memory and wall-clock training time per epoch over a range of time-step sizes. For small $\Delta t$ values, Im-PiNDiff and Ex-PiNDiff exhibit similar memory usage, but the implicit model typically incurs longer training times due to iterative solver overhead. However, Im-PiNDiff supports significantly larger $\Delta t$ while maintaining predictive accuracy. For instance, to achieve a relative error below $0.05\%$, Ex-PiNDiff requires $\Delta t = 1 \times 10^{-4}$ s, whereas Im-PiNDiff remain this accuracy even at $\Delta t = 2 \times 10^{-3}$ s. At this larger step size, the number of rollout steps is reduced by a factor of 20, yielding a 14× reduction in memory usage and a 3× speedup in training time. For most practical problems, the effective inequality $(T/\Delta t_{\text{Im-PiNDiff}})\tilde{K} < (T/\Delta t_{\text{Ex-PiNDiff}})$ always holds, yielding superior performance. In the case of the naive Im-PiNDiff model, the memory usage increases linearly with the specified inner iteration $K$, and for $K=32$, it consumes around 34 GBs. Since training time scales directly with simulation execution time, the findings of training time are equally applicable to inference time.
To further reduce memory requirements, we apply checkpointing to the implicit model (Im-Cp). This technique stores selected intermediate states and recomputes others during the backward pass, significantly reducing memory usage while incurring a modest increase in computational time. As shown in Fig.~\ref{fig:CompCost}, checkpointing flattens the memory growth curve across increasing simulation lengths, making long-horizon training feasible on resource-constrained hardware.

\begin{figure}[!htp]
    \centering
    \includegraphics[width=0.7\linewidth]{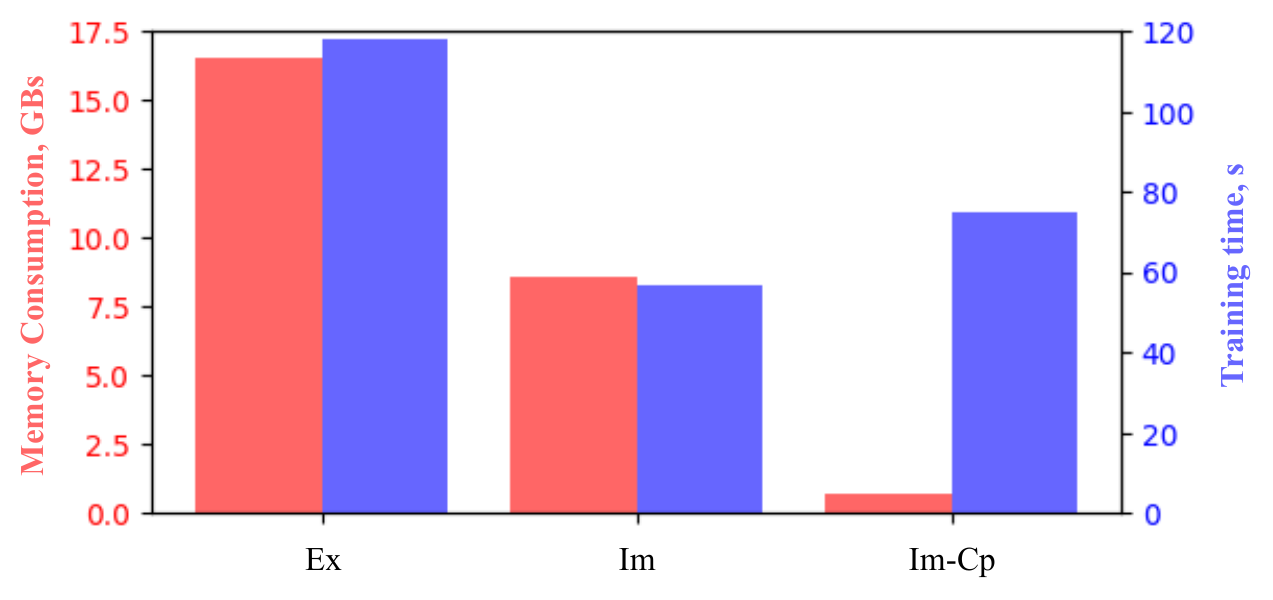} 
    \caption{Comparision of computational cost for Ex-PiNDiff, Im-PiNDiff, and Im-PiNDiff-Cp for CVI modeling.({Ex}: Ex-PiNDiff, {Im}: Im-PiNDiff, {Im-Cp}: Im-PiNDiff w/t Checkpoint) }%
    \label{fig:CompCost_CVI}
\end{figure}

To further demonstrate the scalability and versatility of the hybrid adjoint-based gradient propagation strategy, we investigated a multi-physics reaction–diffusion problem arising in Chemical Vapor Infiltration (CVI) modeling. Specifically, we applied our approach to the PiNDiff-CVI framework developed in~\cite{akhare2024probabilistic}, which simulates porous infiltration by solving two tightly coupled PDEs: an elliptic Poisson equation governing steady-state molarity distribution and a hyperbolic transport equation modeling time-evolving deposition dynamics. The elliptic component introduces an additional layer of complexity, as it necessitates an inner numerical solve at each time step, leading to a nested bilevel optimization structure (More details can be found in \ref{app:CVI}). In the original implementation of PiNDiff-CVI~\cite{akhare2024probabilistic}, the Poisson solver was handled using a fixed number of unrolled iterations, incurring considerable memory overhead during training due to the explicit differentiation of each inner step. In contrast, we replaced this naive backpropagation scheme with our proposed adjoint-based differentiation strategy, which computes gradients through the elliptic solver using the implicit function theorem, thus avoiding the need to store intermediate iterates. For the hyperbolic time integration component, we employed checkpointing to reduce memory requirements by recomputing selected intermediate states during the backward pass. This combination of adjoint differentiation and checkpointing significantly improves both efficiency and scalability. The comparative performance of the original Ex-PiNDiff, Im-PiNDiff, and Im-PiNDiff with checkpointing (Im-Cp) is summarized in Fig.~\ref{fig:CompCost_CVI}. 
The hybrid Im-PiNDiff model achieved nearly a 2× reduction in both memory consumption and wall-clock training time relative to the explicit baseline, without sacrificing predictive accuracy. Moreover, the integration of checkpointing further reduced memory usage to approximately 696 MB, an order of magnitude lower than the Ex-PiNDiff baseline, while incurring only a modest increase in computational time. These results highlight the practicality of hybrid gradient propagation strategies for large-scale, stiff, or tightly coupled PDE systems, where traditional AD approaches are often prohibitive.

\section{Conclusion}
\label{sec:conclusion}
This work presents a hybrid neural-physics modeling framework, termed \textit{Im-PiNDiff}, for complex spatiotemporal dynamics. By introducing implicit neural differential layers into the PiNDiff architecture, we address the challenge of numerical instability and error accumulation inherent in explicit recurrent formulations. The use of implicit time stepping significantly improves the temporal stability and long-horizon predictive accuracy.

A key innovation of the proposed framework lies in its hybrid gradient propagation strategy, which integrates adjoint-based implicit differentiation with reverse-mode AD. This approach decouples gradient computation from the number of solver iterations, enabling memory-efficient training without compromising accuracy. Moreover, we incorporate checkpointing schemes to further reduce the peak memory footprint, making the framework viable for large-scale, long-time simulations on memory-constrained hardware. Together, these algorithmic advances allow Im-PiNDiff to scale to previously intractable problem regimes, achieving superior efficiency and stability over traditional AD-based implementations.

For latent physics inference, we leverage CNFs to parameterize spatially and temporally varying physical quantities, which enables the model to recover unobserved fields or operators from sparse, indirect measurements, expanding the applicability of PiNDiff models to scenarios where direct supervision is unavailable or limited. Extensive numerical experiments, including linear advection-diffusion, nonlinear Burgers' dynamics, and multiphysics CVI processes, demonstrate the proposed framework's effectiveness in both forward and inverse modeling tasks under challenging data and conditions.

Overall, Im-PiNDiff represents a significant step toward enabling stable, efficient, and generalizable hybrid modeling for real-world scientific systems. The combination of implicit architectures, adjoint-based training, and neural field parameterizations offers a flexible and robust paradigm for next-generation scientific machine learning. Future extensions will explore adaptive solvers, stochastic PDEs, and coupling with experimental data streams to further broaden the utility of this framework in data-constrained and multiscale physical modeling settings.

\section*{Declaration of competing interests}
The authors declare no competing interests.



\section*{Data availability}
All data needed to evaluate the conclusions in the paper are either present in the paper or can be regenerated by the code provided.

\section*{Acknowledgment}
The authors would like to acknowledge the funds from the Air Force Office of Scientific Research (AFOSR), United States of America under award number FA9550-22-1-0065. JXW would also like to acknowledge the funding support from the Office of Naval Research under award number N00014-23-1-2071 and the National Science Foundation under award number OAC-2047127 in supporting this study.

\renewcommand*{\bibfont}{\footnotesize}

\clearpage

\appendix

\section{Backpropagation for Autoregressive model}
\label{app:Backpropagation}

\begin{figure}[htp!]
    \centering
    \includegraphics[width=\linewidth]{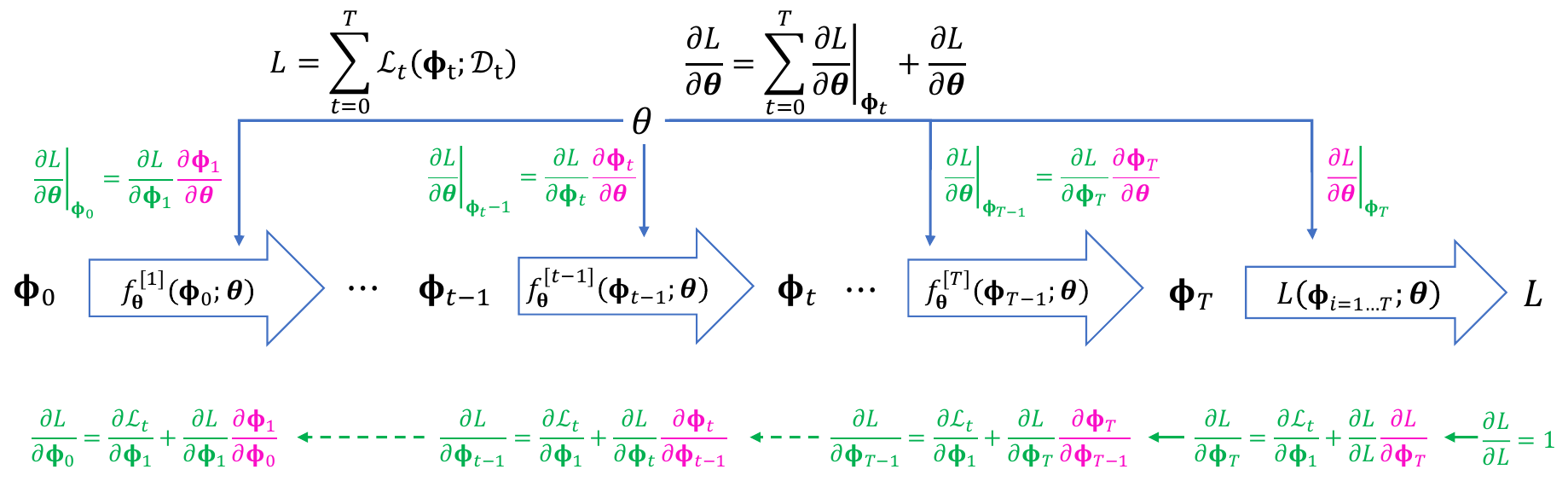}
    \caption{Forward and Back-propagation for Autoregressive model}
    \label{fig:FB-prop}
\end{figure}

Consider a one-step function $\boldsymbol{\Phi}_{t+1} = f_{\boldsymbol{\theta}}(\boldsymbol{\Phi}_{t})$ with $\boldsymbol{\Phi}_{t} \in \mathbb{R}^n$ called at every time step to generated the temporal dynamics. Ex-PiNDiff model's forward computation to generate temporal dynamics up to time $T$, can be expressed as:
\begin{equation}
    \boldsymbol{\Phi}_T = f_{\boldsymbol{\theta}}(\boldsymbol{\Phi}_{T-1}) = f_{\boldsymbol{\theta}}(f_{\boldsymbol{\theta}}(\boldsymbol{\Phi}_{T-2})) = \cdots = f_{\boldsymbol{\theta}}(\cdots f_{\boldsymbol{\theta}}(f_{\boldsymbol{\theta}}(\boldsymbol{\Phi}_{0})))
\end{equation}
or
\begin{equation}
    \boldsymbol{\Phi}_T = f^{[T]}_{\boldsymbol{\theta}}\cdot f^{[T-1]}_{\boldsymbol{\theta}} \cdot f^{[T-2]}_{\boldsymbol{\theta}} \cdots f^{[1]}_{\boldsymbol{\theta}}(\boldsymbol{\Phi}_{0}).
\end{equation}
Let's represent the forward trajectory of states as
\begin{equation}
    \boldsymbol{\Phi}_0 \rightarrow \boldsymbol{\Phi}_1 \cdots \boldsymbol{\Phi}_t \rightarrow \boldsymbol{\Phi}_{t+1} \cdots \boldsymbol{\Phi}_{T-1} \rightarrow \boldsymbol{\Phi}_T\rightarrow L.
\end{equation}
The total loss $L(\boldsymbol{\theta})$ is defined over the entire rollout trajectory of $T$ steps as
\begin{equation}
    L(\boldsymbol{\theta}) = \sum_{t=0}^{T}\mathcal{L}_t(\boldsymbol{\Phi}_t;\mathcal{D}_t) + \mathcal{L}_{regulate}(\boldsymbol{\theta}), 
\end{equation}
and the gradient of loss with respect to $\boldsymbol{\theta}$ can be computed using the chain rule,
\begin{equation}
    \frac{dL}{d\boldsymbol{\theta}} = \frac{\partial L}{
    \partial \boldsymbol{\theta}} + \sum_{t=0}^{T} \frac{\partial L}{
    \partial \boldsymbol{\theta} }\bigg|_{\boldsymbol{\Phi}_t}, 
\end{equation}
where
\begin{subequations}
\begin{align}
    &\frac{\partial L}{
    \partial \boldsymbol{\theta} }\bigg|_{\boldsymbol{\Phi}_t} = \frac{\partial L}{ \partial \boldsymbol{\Phi}_{t+1}} \cdot \Bigg[\frac{\partial \boldsymbol{\Phi}_{t+1}}{ \partial \boldsymbol{\theta}} \Bigg] = \frac{\partial L}{ \partial \boldsymbol{\Phi}_{t+1}} \cdot \Bigg[\frac{\partial f}{ \partial \boldsymbol{\theta}} \Bigg],\\
    &\frac{\partial L}{\partial \boldsymbol{\Phi}_t} = \frac{\partial \mathcal{L}_t}{\partial \boldsymbol{\Phi}_{t}} + \frac{\partial L}{\partial \boldsymbol{\Phi}_{t+1}} \cdot \Bigg[\frac{\partial \boldsymbol{\Phi}_{t+1}}{ \partial \boldsymbol{\Phi}_{t}} \Bigg]  = \frac{\partial \mathcal{L}_t}{\partial \boldsymbol{\Phi}_{t}} + \frac{\partial L}{\partial \boldsymbol{\Phi}_{t+1}} \cdot \Bigg[\frac{\partial f}{ \partial \boldsymbol{\Phi}_{t}} \Bigg],
\end{align}
\end{subequations}
Here $\frac{\partial L}{
    \partial \boldsymbol{\Phi}_{t+1}} \Big[\frac{\partial f}{
    \partial \boldsymbol{\theta}} \Big]$ and $\frac{\partial L}{
    \partial \boldsymbol{\Phi}_{t+1}} \Big[\frac{\partial f}{
    \partial \boldsymbol{\Phi}_{t}} \Big]$ are VJPs, and $[\cdot]$ represent Jacobian matrix.

\section{Gradient backpropagation for implicit-PiNDiff with naive AD}
\label{app:im-PiNDiff}

In case of Im-PiNDiff, we employ a $\text{RootFind}(\cdot)$ method to find the state by finding solution for $f_{\boldsymbol{\theta}}^{im}\big(\boldsymbol{\Phi}_{t}; \boldsymbol{\Phi}_{t-1}^*\big) = 0$ at each time step, which can be expressed as
\begin{equation}
    \boldsymbol{\Phi}_{t}^* \gets \text{RootFind}\Big( f_{\boldsymbol{\theta}}^{im}\big(\boldsymbol{\Phi}_{t}; \boldsymbol{\Phi}_{t-1}^*\big) = 0 \Big), \quad t = 1,\cdots , T.
\end{equation}
or
\begin{equation}
    \boldsymbol{\Phi}_{T}^* \gets \text{RootFind}( f_{\boldsymbol{\theta}}^{im}) \cdot \text{RootFind}( f_{\boldsymbol{\theta}}^{im}) \cdots \text{RootFind}\Big( f_{\boldsymbol{\theta}}^{im}\big(\boldsymbol{\Phi}_{1}; \boldsymbol{\Phi}_{0}\big) = 0 \Big).
\end{equation}
The iterative solver $\text{RootFind}(\cdot)$ creates a sequence of guesses
\begin{equation}
    \{\boldsymbol{\Phi}_{t}^{[0]}, \boldsymbol{\Phi}_{t}^{[1]}, \cdots, \boldsymbol{\Phi}_{t}^{[K]} \} = \Big\{  \text{RootFind}^{[k]}\Big( f_{\boldsymbol{\theta}}^{im}\big(\boldsymbol{\Phi}_{t}^{[0]}; \boldsymbol{\Phi}_{t-1}\big) = 0 \Big) \Big\}_{k=0}^K
\end{equation}
that converge to final solution satisfying $ f_{\boldsymbol{\theta}}^{im}\big(\boldsymbol{\Phi}_{t}^{[K]}; \boldsymbol{\Phi}_{t-1}\big) = 0$.
For Naive AD, we can consider a fixed number of iterations $K$ for $\text{RootFind}(\cdot)$ algorithm. Here, the forward propagation of state looks like
\begin{equation}
    \boldsymbol{\Phi}_0 \rightarrow \big\{\boldsymbol{\Phi}_{1}^{[0]}  \cdots \boldsymbol{\Phi}_{1}^{[K]} \big\}\rightarrow\boldsymbol{\Phi}_{1}^* \cdots \big\{\boldsymbol{\Phi}_{t}^{[0]}  \cdots \boldsymbol{\Phi}_{t}^{[K]} \big\}\rightarrow\boldsymbol{\Phi}_{t}^* \cdots\big\{\boldsymbol{\Phi}_{T}^{[0]} \cdots \boldsymbol{\Phi}_{T}^{[K]} \big\}\rightarrow\boldsymbol{\Phi}_{T}^*
\end{equation}
and the total loss $L(\boldsymbol{\theta})$ defined over the entire rollout trajectory of $T$ steps will be,
\begin{equation}
    L(\boldsymbol{\theta}) = \sum_{t=0}^{T}\mathcal{L}_t(\boldsymbol{\Phi}_t^*;\mathcal{D}_t) + \mathcal{L}_{regulate}(\boldsymbol{\theta}).
\end{equation}
The gradient of loss with respect to $\boldsymbol{\theta}$ for naive Im-PiNDiff is
\begin{equation}
    \frac{dL}{d\boldsymbol{\theta}} = \frac{\partial L}{
    \partial \boldsymbol{\theta}} + \sum_{t=1}^{T} \frac{\partial L}{
    \partial \boldsymbol{\theta} }\bigg|_{\boldsymbol{\Phi}_t^*} = \frac{\partial L}{
    \partial \boldsymbol{\theta}} + \sum_{t=1}^{T} \sum_{k=0}^{K} \frac{\partial L}{
    \partial \boldsymbol{\theta} }\bigg|_{\boldsymbol{\Phi}_t^{[k]}}, 
\end{equation}
where the AD will store every intermediate iterates $\big\{\boldsymbol{\Phi}_{t}^{[0]}  \cdots \boldsymbol{\Phi}_{t}^{[K]} \big\}$ at every time step in addition to storing $\boldsymbol{\Phi}_t^*$ at every time step, resulting in substantial memory overhead. 


\section{Derivation of adjoint-based VJP for implicit layers}
\label{app:adj-VJP-derivation}

To save memory, we need to compute $\frac{\partial L}{ \partial \boldsymbol{\theta} }\big|_{\boldsymbol{\Phi}_t^*}$ without needing to save the intermediate iterates $\big\{\boldsymbol{\Phi}_{t}^{[0]}  \cdots \boldsymbol{\Phi}_{t}^{[K]} \big\}$. Recall $\frac{\partial L}{
    \partial \boldsymbol{\theta} }\big|_{\boldsymbol{\Phi}_t^*} = \frac{\partial L}{ \partial \boldsymbol{\Phi}_{t+1}^*} \cdot \Big[\frac{\partial \boldsymbol{\Phi}_{t+1}^*}{ \partial \boldsymbol{\theta}} \Big],\frac{\partial L}{\partial \boldsymbol{\Phi}_t^*} = \frac{\partial \mathcal{L}_t}{\partial \boldsymbol{\Phi}_{t}^*} + \frac{\partial L}{
    \partial \boldsymbol{\Phi}_{t+1}^*} \cdot \Big[\frac{\partial \boldsymbol{\Phi}_{t+1}^*}{ \partial \boldsymbol{\Phi}_{t}^*} \Big]$.
So we need to compute VJPs: $\frac{\partial L}{
    \partial \boldsymbol{\Phi}_{t+1}^*} \Big[\frac{\partial \boldsymbol{\Phi}_{t+1}^*}{
    \partial \boldsymbol{\theta}} \Big]$ and $\frac{\partial L}{
    \partial \boldsymbol{\Phi}_{t+1}^*} \Big[\frac{\partial \boldsymbol{\Phi}_{t+1}^*}{
\partial \boldsymbol{\Phi}_{t}^*} \Big]$ for the implicit layer. The joint-state method provides a way to compute these VJPs as a solution of the linear system. The derivation of the linear equation for obtaining VJPs is obtained using the implicit function theorem.

\textbf{Implicit function theorem}:\cite{blondel2024elements}

For a function $f(\boldsymbol\omega, \boldsymbol\lambda): W\times\Lambda\rightarrow W$, that is continuously differentiable function in a neighborhood of $(\boldsymbol\omega_0, \boldsymbol\lambda_0)$ with $f(\boldsymbol\omega, \boldsymbol\lambda) = \textbf{0}$ and $\frac{\partial}{\partial \boldsymbol\omega} f(\boldsymbol\omega, \boldsymbol\lambda)$ is invertible, then there exists a neighbourhood of $\boldsymbol\lambda_0$ in which there is a function $\boldsymbol\omega^*(\boldsymbol\lambda)$ such that
\begin{itemize}
    \item $\boldsymbol\omega^*(\boldsymbol\lambda) = \boldsymbol\omega_0$
    \item $f(\boldsymbol\omega^*(\boldsymbol\lambda), \boldsymbol\lambda) = 0$ for all $\boldsymbol\lambda$ in the neighborhood,
    \item $\frac{\partial }{\partial \boldsymbol\lambda}\boldsymbol\omega^*(\boldsymbol\lambda) = -\bigg[\frac{\partial}{\partial \boldsymbol\omega} f(\boldsymbol\omega^*(\boldsymbol\lambda), \boldsymbol\lambda)\bigg]^{-1} \frac{\partial}{\partial \boldsymbol\lambda} f(\boldsymbol\omega^*(\boldsymbol\lambda), \boldsymbol\lambda)$
\end{itemize}

Conisder $f^{im}\big(\boldsymbol{\Phi}_{t}; \boldsymbol{\Phi}_{t-1}^*, \boldsymbol{\theta}\big)$ with $\boldsymbol{\Phi}_{t-1}^*, \boldsymbol{\theta}$ as $\boldsymbol\lambda$, the implicit function theorem provides 
\begin{subequations}
\begin{align}
    &\frac{\partial \boldsymbol{\Phi}_{t+1}^*}{ \partial \boldsymbol{\theta}} = -\left[\frac{\partial f^{im}}{\partial \boldsymbol{\Phi}_{t+1}^*}\right]^{-1} \cdot \frac{\partial f^{im}}{\partial \boldsymbol{\theta}},\\
    &\frac{\partial \boldsymbol{\Phi}_{t+1}^*}{ \partial \boldsymbol{\Phi}_{t}^*} = -\left[\frac{\partial f^{im}}{\partial \boldsymbol{\Phi}_{t+1}^*}\right]^{-1} \cdot \frac{\partial f^{im}}{\partial \boldsymbol{\Phi}_{t}^*}.
\end{align}
\end{subequations}
Multilpy row vector $\frac{\partial L}{\partial \boldsymbol{\Phi}_{t+1}^*}$ on left side, we get
\begin{subequations}
\begin{align}
    &\frac{\partial L}{ \partial \boldsymbol{\Phi}_{t+1}^*} \cdot \Bigg[\frac{\partial \boldsymbol{\Phi}_{t+1}^*}{ \partial \boldsymbol{\theta}} \Bigg] = -\frac{\partial L}{ \partial \boldsymbol{\Phi}_{t+1}^*} \cdot 
 \left[\frac{\partial f^{im}}{\partial \boldsymbol{\Phi}_{t+1}^*}\right]^{-1} \cdot \frac{\partial f^{im}}{\partial \boldsymbol{\theta}},\\
    &\frac{\partial L}{ \partial \boldsymbol{\Phi}_{t+1}^*} \cdot \Bigg[\frac{\partial \boldsymbol{\Phi}_{t+1}^*}{ \partial \boldsymbol{\Phi}_{t}^*} \Bigg] = - \frac{\partial L}{ \partial \boldsymbol{\Phi}_{t+1}^*} \cdot 
 \left[\frac{\partial f^{im}}{\partial \boldsymbol{\Phi}_{t+1}^*}\right]^{-1} \cdot \frac{\partial f^{im}}{\partial \boldsymbol{\Phi}_{t}^*}.
\end{align}
\end{subequations}
Now we define the adjoint vector $\mathbf{w}^T_{t+1} \in \mathbb{R}^{1\times n}$ as,
\begin{equation}
    \mathbf{w}^T = -\frac{\partial L}{  \partial \boldsymbol{\Phi}_{t+1}^*} \cdot \left[\frac{\partial f^{im}}{  \partial \boldsymbol{\Phi}_{t+1}^*} \right]^{-1},
\end{equation}
which is obtained by solving the linear system,
    \begin{equation}
        \textbf{w}^T\bigg[\frac{\partial f^{im}}{  \partial \boldsymbol{\Phi}_{t+1}^*} \bigg]=-\frac{\partial L}{  \partial \boldsymbol{\Phi}_{t+1}^*}.
    \end{equation}
This linear equation is solved efficiently using iterative numerical linear solvers (such as GMRES or conjugate gradient methods)
Finally, the VJPs are expressed as,
\begin{subequations}\label{eq:implicitDiff}
\begin{align}
    &\frac{\partial L}{ \partial \boldsymbol{\Phi}_{t+1}^*} \cdot \Bigg[\frac{\partial \boldsymbol{\Phi}_{t+1}^*}{ \partial \boldsymbol{\theta}} \Bigg] = \mathbf{w}^T_{t+1} \cdot \frac{\partial f^{im}}{\partial \boldsymbol{\theta}},\\
    &\frac{\partial L}{ \partial \boldsymbol{\Phi}_{t+1}^*} \cdot \Bigg[\frac{\partial \boldsymbol{\Phi}_{t+1}^*}{ \partial \boldsymbol{\Phi}_{t}^*} \Bigg] = \mathbf{w}^T_{t+1} \cdot \frac{\partial f^{im}}{\partial \boldsymbol{\Phi}_{t}^*},
\end{align}
\end{subequations}
We can compute $\frac{\partial L}{ \partial \boldsymbol{\theta} }\big|_{\boldsymbol{\Phi}_t^*}$ for every implicit layer without needing to save the intermediate iterates, thereby reducing the memory requirement for the AD backpropagation.

\section{Spatiotemporal fields using a combination of cosine functions}
\label{sec:sinfu}

In order to make inference challenging, a combination of cosine functions is superimposed to generate a more complex spatio-temporal field for advection and viscosity.
\begin{equation}
    f(\mathbf x,t) = \sum_{i=1}^{nw} A_i \sin(2\pi(k_i \mathbf x+\omega_it) + p_i)    
\end{equation}
where $A_i k_i, \omega_i, p_i \sim \mathcal{U}(0,2)$ are randomly sampled.
 $A_i \sim \mathcal{U}(-2,2), k_i \sim \mathcal{U}(0,2), \omega_i \sim \mathcal{U}(-1,1), p_i \sim \mathcal{U}(0,2)$

\section{PiNDiff modeling for chemical vapor infiltration}
\label{app:CVI}
Here we talk in more detail about the reaction-diffusion system, chemical vapor infiltration (CVI) used in this study. CVI is a materials processing technique used to fabricate composite materials. In this method, a porous preform—usually made of fibers like carbon is exposed to a reactive gas mixture at elevated temperatures inside a furnace. The gaseous precursors diffuse into the pores of the preform and undergo chemical reactions, typically decomposition or reduction, to deposit a solid material onto the internal surfaces of the preform. Over time, this gradual deposition fills the pores and forms a dense matrix around the fibers without significantly disturbing the structure. CVI allows for precise control over material composition and microstructure, making it suitable for producing high-temperature, corrosion-resistant components used in aerospace, energy, and defense applications. A PiNDiff-CVI model based on foundational physics was developed to simulate the CVI\cite{akhare2024probabilistic} process, whose equations are given as 
\begin{subequations}
  \begin{equation}
    D_{\text {eff}} \nabla^2( C) = KS_{\text v}C,
  \end{equation}
  \begin{equation}
    \rho_{\text s}\frac{d\varepsilon}{dt} = - qM_{\text s} KS_{\text v}C.
  \end{equation}
  \label{eq:I-CVI}
\end{subequations}
In the above equations, $C = C(\mathbf{x}, t)$ denotes the effective molarity field (mol m$^{-3}$) of all reactive gases, $\varepsilon = \varepsilon(\mathbf{x}, t)$ is the porosity of the preform, $q$ represent a constant stichometric coefficient, $M_d$ is the molar mass (kg mol$^{-1}$), and $\rho_d$ is the density (kg m$^{-3}$) of the deposited solid (carbon or SiC). $D_{eff} = D_{eff}(\mathbf{x}, t)$ represents the effective diffusion coefficient field, $K = K(\mathbf{x}, t)$ is the deposition reaction rate, and $S_v = S_v(\mathbf{x}, t)$ corresponds to the surface-to-volume ratio. 

\begin{figure}
    \centering
    \includegraphics[width=0.7\linewidth]{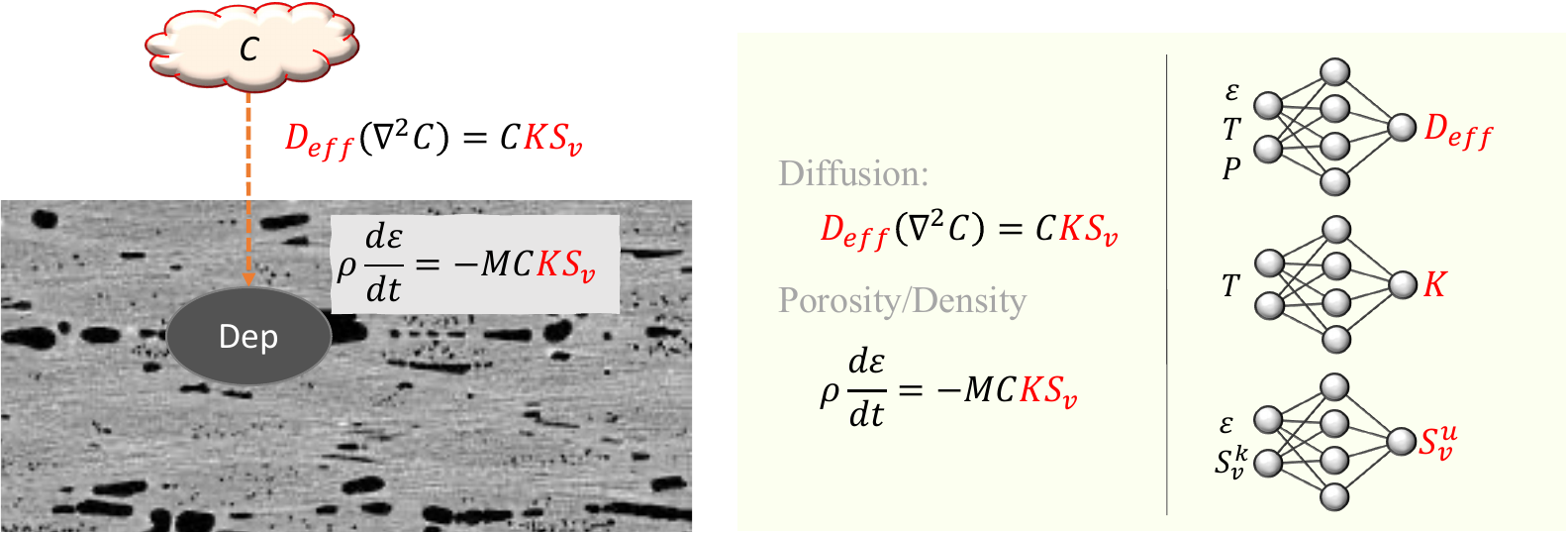}
    \caption{Foundational physics of the CVI process with neural operator approximations}
    \label{fig:cvi}
\end{figure}
The details regarding this solver are described by Akhare et al.\cite{akhare2024probabilistic}. In the PiNDiff-CVI model, the underlying transport and reaction functions were unknown and modeled as the operators $D_{eff}$, $K$, and $S_v$ using DNNs as shown in Fig.~\ref{fig:cvi}. Equation~\ref{eq:I-CVI}a is an elliptic Poisson equation governing steady-state molarity distribution, solved iteratively at each time step using the point-Jacobi method (inner optimization). While Equation~\ref{eq:I-CVI}b is a hyperbolic transport equation modeling time-evolving deposition dynamics, solved explicitly using the Euler time integration scheme. Previously, a naive approach involved using a fixed number of iterations for solving Equation~\ref{eq:I-CVI}a was employed, necessary for constructing a static computational graph required for gradient calculations, resulting in large memory usage. By leveraging adjoint-based backpropagation, we eliminate the need to save the intermediate iterates, thereby reducing memory usage.

\end{document}